\documentclass[runningheads]{llncs}
\usepackage{graphicx}
\usepackage{amsmath,amssymb} % define this before the line numbering.
\usepackage{color}
\usepackage{multirow}
\usepackage{rotating}
\usepackage{booktabs}
\usepackage{multirow}
\usepackage{gensymb}
\usepackage{algorithm}
\usepackage[resetlabels,labeled]{multibib}
\usepackage[pagebackref=true,breaklinks=true,letterpaper=true,colorlinks,bookmarks=false]{hyperref}
\usepackage[width=122mm,left=12mm,paperwidth=146mm,height=193mm,top=12mm,paperheight=217mm]{geometry}

\newcites{S}{References}

\begin{document}

% \linenumbers
\pagestyle{headings}
\mainmatter

\title{Joint Learning of Intrinsic Images and Semantic Segmentation} 
% Replace with your title

\titlerunning{Joint Learning of Intrinsic Images and Semantic Segmentation}
% Replace with a meaningful short version of your title

\authorrunning{Baslamisli et al.}
% Replace with shorter version of the author list. If there are more authors than fits a line, please use A. Author et al.

\author{Anil S. Baslamisli\textsuperscript{1}, Thomas T. Groenestege\textsuperscript{1,2}, Partha Das\textsuperscript{1,2}, Hoang-An Le\textsuperscript{1}, Sezer Karaoglu\textsuperscript{1,2}, Theo Gevers\textsuperscript{1,2}}

%Please write out author names in full in the paper, i.e. full given and family names. 
%If any authors have names that can be parsed into FirstName LastName in multiple ways, please include the correct parsing, in a comment to the volume editors:
%\index{Lastnames, Firstnames}
%(Do not uncomment it, because you may introduce extra index items if you do that, we will use scripts for introducing index entries...)

\institute{\textsuperscript{1}University of Amsterdam, \textsuperscript{2}3DUniversum B.V.\\
	\email{ \{a.s.baslamisli, h.a.le, th.gevers\}@uva.nl},  \email{s.karaoglu@3duniversum.com}
}

\maketitle

\begin{abstract}
Semantic segmentation of outdoor scenes is problematic when there are variations in imaging conditions. It is known that albedo (reflectance) is invariant to all kinds of illumination effects. Thus, using reflectance images for semantic segmentation task can be favorable. Additionally, not only segmentation may benefit from reflectance, but also segmentation may be useful for reflectance computation. Therefore, in this paper, the tasks of semantic segmentation and intrinsic image decomposition are considered as a combined process by exploring their mutual relationship in a joint fashion. To that end, we propose a supervised end-to-end CNN architecture to jointly learn intrinsic image decomposition and semantic segmentation. We analyze the gains of addressing those two problems jointly. Moreover, new cascade CNN architectures for intrinsic-for-segmentation and segmentation-for-intrinsic are proposed as single tasks. Furthermore, a dataset of 35K synthetic images of natural environments is created with corresponding albedo and shading (intrinsics), as well as semantic labels (segmentation) assigned to each object/scene. The experiments show that joint learning of intrinsic image decomposition and semantic segmentation is beneficial for both tasks for natural scenes. Dataset and models are available at:  \url{https://ivi.fnwi.uva.nl/cv/intrinseg}.

%\keywords{Intrinsic image decomposition, semantic segmentation, joint learning, scene understanding}
\end{abstract}

\section{Introduction}
Semantic segmentation of outdoor scenes is a challenging problem in computer vision. Variations in imaging conditions may negatively influence the segmentation process. These varying conditions include shading, shadows, inter-reflections, illuminant color and its intensity. As image segmentation is the process of identifying and semantically grouping pixels, drastic changes in pixel values may hinder a successful segmentation. To address this problem, several methods are proposed to mitigate the effects of illumination to obtain more robust image features to help semantic segmentation \cite{upcroft,wang,suh,ramakrishnan}. Unfortunately, these methods provide illumination invariance artificially by hand crafted features. Instead of using narrow and specific invariant features, in this paper, we focus on image formation invariance induced by a full intrinsic image decomposition.

Intrinsic image decomposition is the process of decomposing an image into its image formation components such as albedo (reflectance) and shading (illumination) \cite{land}. The reflectance component contains the true color of objects in a scene. In fact, albedo is invariant to illumination, while the shading component heavily depends on object geometry and illumination conditions in a scene. As a result, using reflectance images for semantic segmentation task can be favorable, as they do not contain any illumination effect. Additionally, not only segmentation may benefit from reflectance, but also segmentation may be useful for reflectance computation. Information about an object reveals strong priors about its intrinsic properties. Each object label constrains the color distribution and is expected to reflect that property to class specific reflectance values. Therefore, distinct object labels provided by semantic segmentation can guide intrinsic image decomposition process by yielding object specific color distributions per label. Furthermore, semantic segmentation process can act as an object boundary guidance map for intrinsic image decomposition by enhancing cues that differentiate between reflectance and occlusion edges in a scene. In addition, homogeneous regions (i.e. in terms of color) within an object segment should have similar reflectance values. Therefore, in this paper, the tasks of semantic segmentation and intrinsic image decomposition are considered as a combined process by exploring their mutual relationship in a joint fashion. 

To this end, we propose a supervised end-to-end convolutional neural network (CNN) architecture to jointly learn intrinsic image decomposition {\em and} semantic segmentation. The joint learning includes an end-to-end trainable encoder-decoder CNN with one shared encoder and three separate decoders: one for reflectance prediction, one for shading prediction, and one for semantic segmentation prediction. In addition to joint learning, we explore new cascade CNN architectures to use reflectance to improve semantic segmentation, and semantic segmentation to steer the process of intrinsic image decomposition. 

To train the proposed supervised network, a large dataset is needed with ground-truth images for both image semantic segmentation (i.e. class labels) and intrinsic properties (i.e. reflectance and shading). However, there is no such a dataset. Therefore, we have created a large-scale dataset featuring plants and objects under varying illumination conditions that are mostly found in natural environments. The dataset is at scene-level considering natural environments containing intrinsic image decomposition and semantic segmentation ground-truths. The dataset contains 35K synthetic images with corresponding albedo and shading (intrinsics), as well as semantic labels (segmentation) assigned to each object/scene. 

Our contributions are: (1) a CNN architecture for joint learning of intrinsic image decomposition and semantic segmentation, (2) analysis on the gains of addressing those two problems jointly, (3) new cascade CNN architectures for intrinsic-for-segmentation and segmentation-for-intrinsic, and (4) a very large-scale dataset of synthetic images of natural environments with scene level intrinsic image decomposition and semantic segmentation ground-truths.

\section{Related Work}
\textbf{Intrinsic Image Decomposition.} 
Intrinsic image decomposition is an ill-posed and under-constrained problem since an infinite number of combinations of photometric and geometric properties of a scene can produce the same 2D image. Therefore, most of the work on intrinsic image decomposition considers priors about scene characteristics to constrain a pixel-wise optimization task. For instance, both \cite{shen} and \cite{zhao} use non-local texture cues, whereas \cite{gehler} and \cite{shen2} constrain the problem with the assumption of sparsity of reflectance. In addition, the use of multiple images helps to resolve the ambiguity where the reflectance is constant and the illumination changes \cite{weiss,matsushita}. Nonetheless, with the success of supervised deep CNNs \cite{vggnet,rcnn}, more recent research on intrinsic image decomposition has shifted towards using deep learning. \cite{narihia} is the first work that uses end-to-end trained CNNs to address the problem. They argue that the model should learn both local and global cues together with a multi-scale architecture. In addition, \cite{shi} proposes a model by introducing inter-links between decoder modules, based on the expectation that intrinsic components are correlated. Moreover, \cite{lettry} demonstrates the capability of generative adversarial networks for the task. On the other hand, in more recent work, \cite{baslamisli} considers an image formation loss together with gradient supervision to steer the learning process to achieve more vivid colors and sharper edges. 

In contrast, our proposed method jointly learns intrinsic properties and segmentation. Additionally, the success of supervised deep CNNs not only depends on a successful model, but also on the availability of annotated data. Generating ground-truth intrinsic images is only possible in a fully-controlled setup and it requires enormous effort and time \cite{mit}. To that end, the most popular real-world dataset for intrinsic image decomposition includes only 20 object-centered images with their ground-truth intrinsics \cite{mit}, which alone is not feasible for deep learning. On the other hand, \cite{iiw} presents scene-level real world relative reflectance comparisons over point pairs of indoor scenes. However, it does not include ground-truth intrinsic images. The most frequently used scene-level synthetic dataset for intrinsic image decomposition is the MPI Sintel Dataset \cite{sintel}. It provides around a thousand of cartoon-like images with their ground-truth intrinsics. 
Therefore, a new dataset is created consisting of 35K synthetic (outdoor) images with 16 distinct object types/scenes which are recorded under different illumination conditions. The dataset contains intrinsic properties and object segmentation ground-truth labels. The dataset is described in detail in the experimental section.

\noindent \textbf{Semantic Segmentation.}
Traditional semantic segmentation methods design hand-crated features to achieve per-pixel classification with the use of an external classifier such as support vector machines \cite{fulkerson,csurka,shotton}. On the other hand, contemporary semantic segmentation methods such as \cite{segnet,fcnn,deeplab} benefit from the powerful CNN models and large-scale datasets such as \cite{pascal,cityscape}. A detailed review on deep learning techniques applied to semantic segmentation task can be found in \cite{garcia}. 

Photometric changes, which are due to varying illumination conditions, cause changes in the appearance of objects. Consequently, these appearance changes create problems for the semantic segmentation task. Therefore, several methods are proposed to mitigate the effects of varying illumination to accomplish a more robust semantic segmentation by incorporating illumination invariance in their algorithms \cite{upcroft,wang,suh,ramakrishnan}. However, these methods provide invariance artificially by hand crafted features. Therefore, they are limited in compensating for possible changes in photometry (i.e. illumination). Deep learning based methods may learn to accommodate photometric changes through data exploration. However, they are constrained by the amount of data. In this paper, we propose to use the intrinsic reflectance property (i.e. fully illumination invariance) to be used for semantic segmentation.

\noindent \textbf{Joint Learning.} Semantic segmentation has been used for joint learning tasks as it provides useful cues about objects and scenes. For instance, \cite{jafari,eigen,mousavian} propose joint depth prediction and semantic segmentation models. Joint semantic segmentation and 3D scene reconstruction is proposed by \cite{kundu}. Furthermore, \cite{ladicky} formulates dense stereo reconstruction and semantic segmentation in a joint framework.

For intrinsic image decomposition, \cite{barron} introduces the first unified model for recovering shape, reflectance, and chromatic illumination in a joint optimization framework. Other works \cite{kim2,shelhamer}, jointly predict depth and intrinsic property. Finally, \cite{vineet} exploits the relation between the intrinsic property and objects (i.e. attributes and segments). The authors propose to address these problems in a joint optimization framework. Using hand crafted priors, \cite{vineet} designs energy terms per component and combines them in one global energy to be minimized. In contrast to previous methods, our proposed method is an end-to-end solution and does not rely on any hand crafted priors. Additionally, \cite{vineet} does not optimize their energy function for each component separately. Therefore, the analysis on the influence of intrinsic image decomposition on semantic segmentation is omitted. In this paper, an in-depth analysis for each component is given.

\section{Approach}

\subsection{Image Formation Model}
To formulate our intrinsic image decomposition, the diffuse reflectance component is considered \cite{shafer}. Then, an $RGB$ image, \textit{I}, over the visible spectrum $\omega$, is defined by:

\begin{equation} \label{eq:brdf2}
\begin{aligned}
I = m_b(\vec{n}, \vec{s}) \int_{\omega}^{} f_{c}(\lambda)\;e(\lambda)\;\rho_{b}(\lambda)\; \mathrm{d}\lambda .
\end{aligned}
\end{equation}

\noindent In the equation, $\vec{n}$ denotes the surface normal, whereas $\vec{s}$ is the light source direction; together forming the geometric dependencies \textit{m}, which in return forms the shading component $S(\vec{x})$ under white light. Additionally, $\lambda$ represents the wavelength, $f_{c}(\lambda)$ is the camera spectral sensitivity, $e(\lambda)$ specifies the spectral power distribution of the illuminant, and $\rho_{b}$ represents the diffuse surface reflectance $R(\vec{x})$. Then, using narrow band filters and considering a linear sensor response under white light, intrinsic image decomposition can be formulated as:

\begin{equation} \label{eq:iid}
I(\vec{x}) = R(\vec{x}) \times S(\vec{x}) .
\end{equation}

\noindent Then, for a position $\vec{x}$, $I(\vec{x})$ can be approximated by the element-wise product of its intrinsic components. When the light source is colored, it is also  included in the shading component.

\begin{figure*}[t]
\includegraphics[scale=0.35]{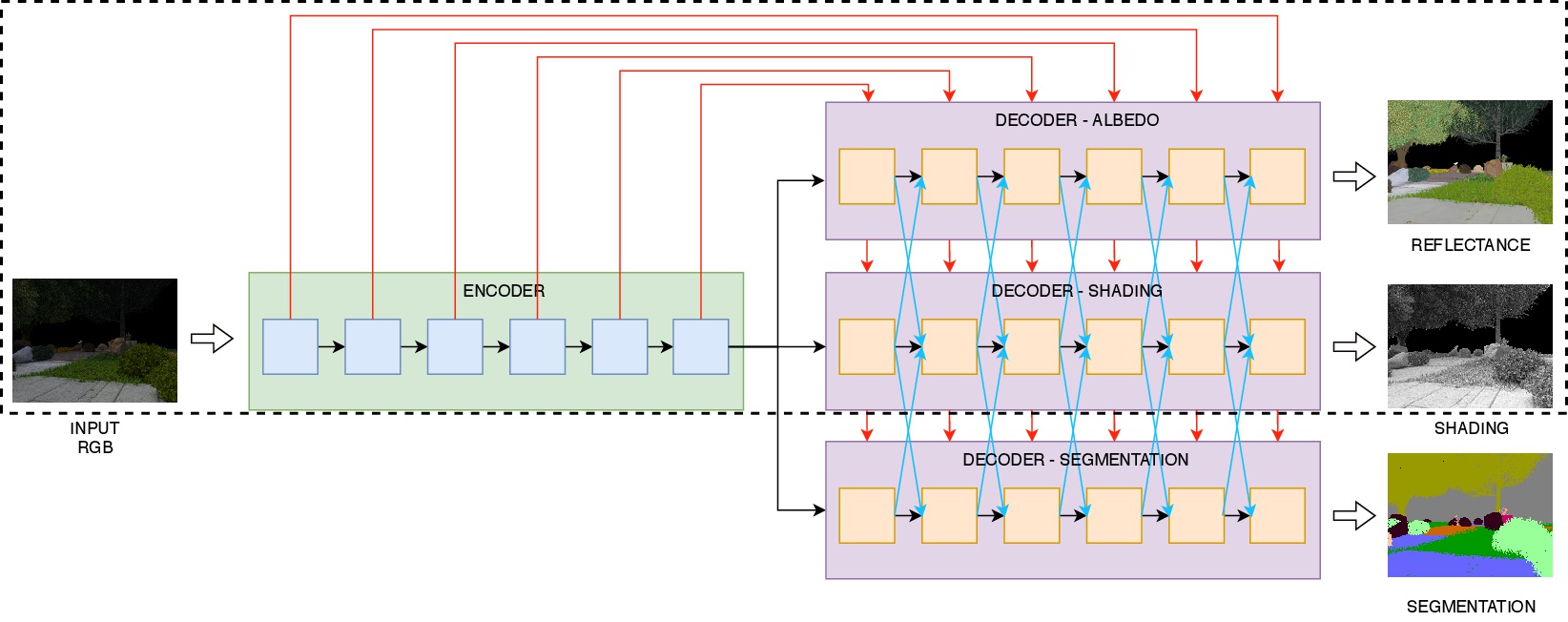}
\centering
\caption{Model architecture for jointly solving intrinsic image decomposition and semantic segmentation with one shared encoder and three separate decoders:  one for shading, one
for reflectance, and one for semantic segmentation prediction. The part in the dotted rectangle denotes the baseline ShapeNet model of \cite{shi}.}
\label{fig:joint_model}
\end{figure*}

\subsection{Baseline Model Architectures}
\textbf{Intrinsic Image Decomposition.}  We use the model proposed by \cite{shi}, $ShapeNet$, without the specular highlight module. The model is shown in the dotted rectangle part of Figure \ref{fig:joint_model}. The model provides state-of-the results for intrinsic image decomposition task. Early features in the encoder block are connected with the corresponding decoder layers, which are called \textit{mirror links}. That proves to be useful for keeping visual details and producing sharp outputs. Furthermore, the features across the decoders are linked to each other (\textit{inter-connections}) to further strengthen the correlation between the components.

To train the model for intrinsic image decomposition task, we use a combination of the standard $L_2$ reconstruction loss (MSE) with its scale invariant version (SMSE). Let $J$ be the prediction of the network and $\hat{J}$ be the ground-truth intrinsic image. Then, the standard $L_2$ reconstruction loss $\mathcal{L}_{MSE}$ is given by:

\begin{equation} \label{eq:MSE}
\mathcal{L}_{MSE}(J, \hat{J}) = \frac{1}{n} \sum_{\vec{x},c}^{} ||\hat{J} - J||^{2}_{2},
\end{equation}

\noindent where $\vec{x}$ denotes the pixel coordinate, $c$ is the color channel index and $n$ is the total number of evaluated pixels. Then, SMSE scales $J$ first and compares MSE with $\hat{J}$:

\begin{equation} \label{eq:SMSE}
\mathcal{L}_{SMSE}(J, \hat{J}) = \mathcal{L}_{MSE}(\alpha J, \hat{J}),
\end{equation}

\begin{equation} \label{eq:alpha}
\alpha =  argmin \; \mathcal{L}_{MSE}(\alpha J, \hat{J}).
\end{equation}

\noindent Then, the combined loss $\mathcal{L}_{CL}$ for training an intrinsic component becomes:

\begin{equation} \label{eq:CL}
\begin{aligned}
\mathcal{L}_{CL}(J, \hat{J}) = \gamma_{SMSE} \; \mathcal{L}_{SMSE}(J, \hat{J}) + \gamma_{MSE} \; \mathcal{L}_{MSE}(J, \hat{J}),
\end{aligned}
\end{equation}

\noindent where the $\gamma$s are the corresponding loss weights. The final loss $\mathcal{L}_{IL}$ for training the model for intrinsic image decomposition task becomes:

\begin{equation} \label{eq:CL2}
\begin{aligned}
\mathcal{L}_{IL}(R, \hat{R}, S, \hat{S}) = \gamma_{R} \; \mathcal{L}_{CL}(R, \hat{R}) + \gamma_{S} \; \mathcal{L}_{CL}(S, \hat{S}).
\end{aligned}
\end{equation}

\noindent \textbf{Semantic segmentation} The same architecture is used as the baseline for semantic segmentation task. However, one of the decoders is removed from the architecture, because there is only one task. As a consequence, inter-connection links are not used for the semantic segmentation task. Furthermore, as a second baseline, we train an off-the-shelf segmentation algorithm~\cite{segnet}, $SegNet$, that is specifically engineered for semantic segmentation task.  

To train the model for semantic segmentation, we use the cross entropy loss:

\begin{equation} \label{eq:ce}
\mathcal{L}_{CE} = - \frac{1}{n} \sum_{\vec{x}}^{} \sum_{L \in O_{\vec{x}}}^{} \;log(p_{\vec{x}}^{L}) \; ,
\end{equation}

\noindent where $p$ is the output of the softmax function to compute the posterior probability of a given pixel $\vec{x}$ belonging to $L^{th}$ class, where $L \in O_{\vec{x}}$ and $O_{\vec{x}} = \{0, 1, 2, \cdot \cdot \cdot , C\}$ as the category set for pixel level class label.

\subsection{Joint Model Architecture}
In this section, a new joint model architecture is proposed. It is an extension of the base model architecture for intrinsic image decomposition task, $ShapeNet$~\cite{shi}, that combines the two tasks i.e. intrinsic image decomposition and semantic segmentation. We modify the baseline model architecture to have one encoder and three distinct decoders i.e. one for reflectance prediction, one for shading prediction, and one for semantic segmentation prediction. We maintain the mirror links and inter-connections. That allows for the network to be constrained with different outputs, and thus reinforce the learned features from different tasks. As a result, the network is forced to learn joint features for the two tasks at hand not only in the encoding phase, but also in the decoding phase. Both encoder and decoder parts contain both intrinsic properties and semantic segmentation characteristics. This setup is expected to be exploited by individual decoder blocks to learn extra cues for the task at hand. Figure \ref{fig:joint_model} illustrates the joint model architecture. To train the model jointly, we combine the task specific loss functions by summing them together:

\begin{equation} \label{eq:jointL}
\begin{aligned}
\mathcal{L}_{JL}(I, R, \hat{R}, S, \hat{S}) = \gamma_{CE} \; \mathcal{L}_{CE} + \gamma_{IL} \; \mathcal{L}_{IL}(R, \hat{R}, S, \hat{S}).
\end{aligned}
\end{equation}

\noindent The effect of the gamma parameters of Equation~\ref{eq:CL} and more implementation details can be found in the supplementary materials.

\section{Experiments}
\subsection{New Synthetic Dataset of Natural Environments}

\begin{figure}[t]
    \centering
    \includegraphics[width=.8\textwidth]{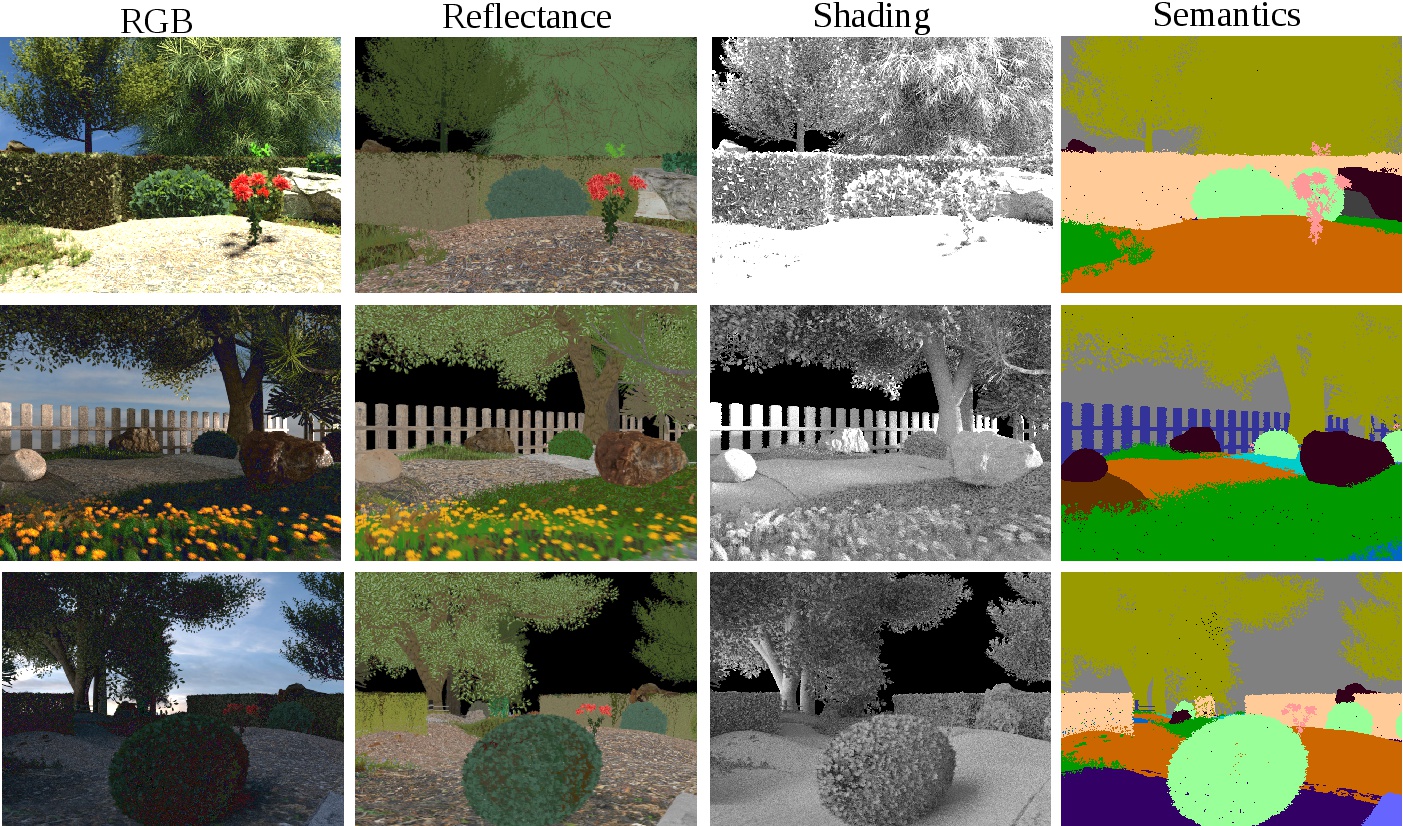}
    \caption{Sample images from the Natural Environment Dataset (NED) featuring plants and objects under varying illumination conditions with ground-truth components}
    \label{fig:garden}
\end{figure}

A large set of synthetic images is created featuring plants and objects that are mostly found in natural environments such as parks and gardens. The dataset contains different species of vegetation such as trees and flowering plants with different types of terrains and landscapes under different lighting conditions. Furthermore, scenarios are created which involves human intervention such as the presence of bushes (like rectangular hedges or spherical topiaries), fences, flowerpots and planters, and etc. (16 classes in total). There is a substantial variety of object colors and geometry. The dataset is constructed by using the parametric tree models \cite{Weber} (implemented as 
add-ons in Blender software), and several manually-designed models from the Internet that aim for 
realistic natural scenes and environments. Ambient lighting is provided by real HDR sky images with a parallel light source. Light source properties are designed to correspond to daytime lighting conditions such as clear sky, cloudy, sunset, twilight, etc. For each virtual park/garden, we captured the scene from different perspectives with motion blur effects. Scene are rendered with the physics-based Blender Cycles\footnote{\url{https://www.blender.org/}} 
engine. To obtain annotations, the rendering pipeline is modified to output $RGB$  images, their corresponding albedo and shading profiles (intrinsics) and semantic labels (segmentation). The dataset consists of 35K images, depicted 40 various parks/gardens under 5 lighting conditions. A number of  samples are shown in Figure \ref{fig:garden}. For the experiments, the dataset is randomly split into 80\% training and 20\% testing (scene split).

\subsection{Error Metrics}
To evaluate our method for intrinsic image decomposition task, we report on mean squared error (MSE), its scale invariant version (SMSE), local mean squared error (LMSE), and dissimilarity version of the structural similarity index (DSSIM). DSSIM accounts for the perceptual visual quality of the results. Following \cite{mit}, for MSE, the absolute brightness of each image is adjusted to minimize the error. Further, $k$ = 20 is used for the window size of LMSE. For semantic segmentation task, we report on global pixel accuracy, mean class accuracy and mean intersection over union (mIoU).

\section{Evaluation}

\subsection{Influence of Reflectance on Semantic Segmentation}
\begin{figure}[t]
    \centering
    \includegraphics[width=.8\textwidth]{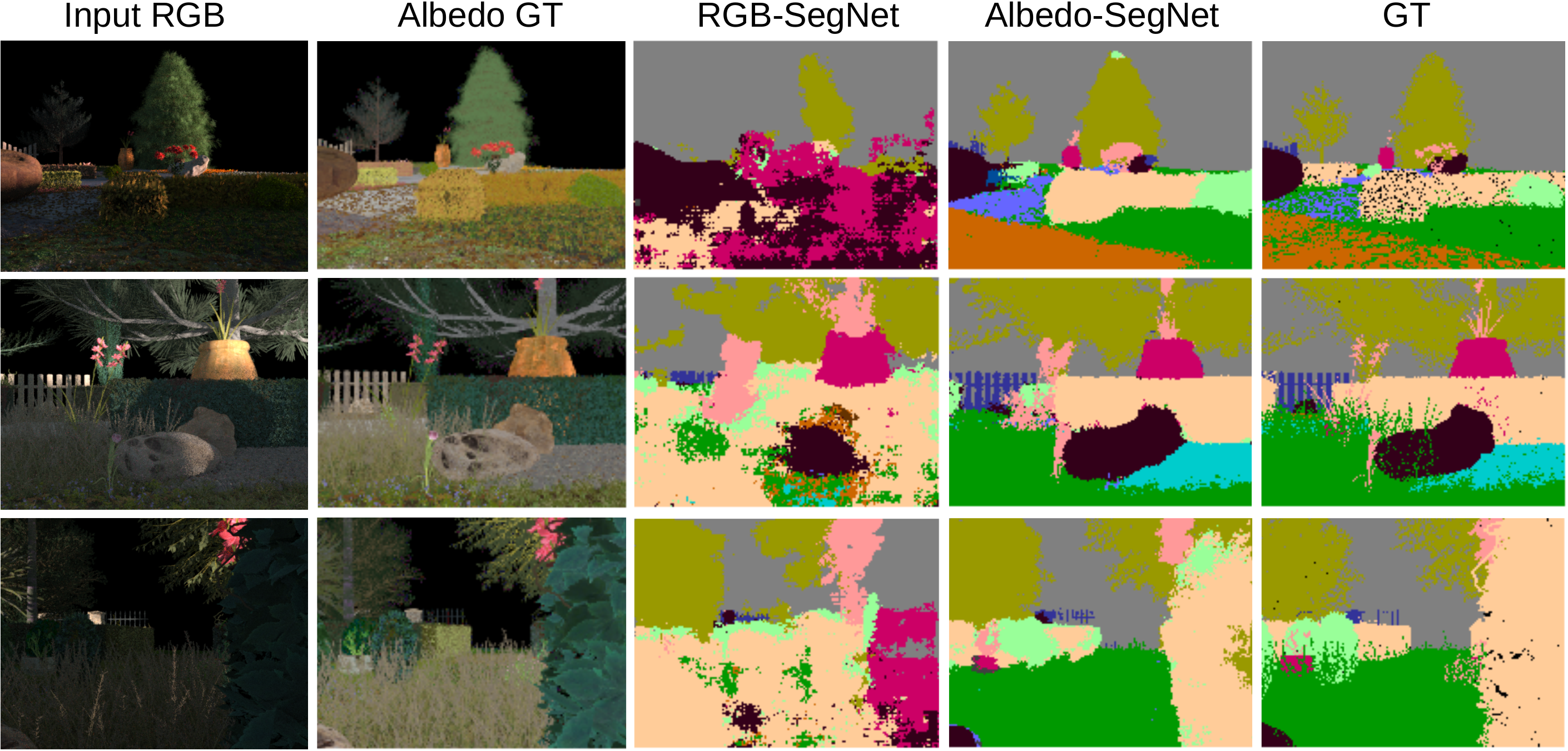}
    \caption{Qualitative evaluation of the influence of reflectance on semantic segmentation. The results show that the semantic segmentation algorithm highly benefits from illumination invariant intrinsic properties (i.e. reflectance)}
    \label{fig:segnet2}
\end{figure}

\begin{figure}[t]
    \centering
    \includegraphics[width=.75\textwidth]{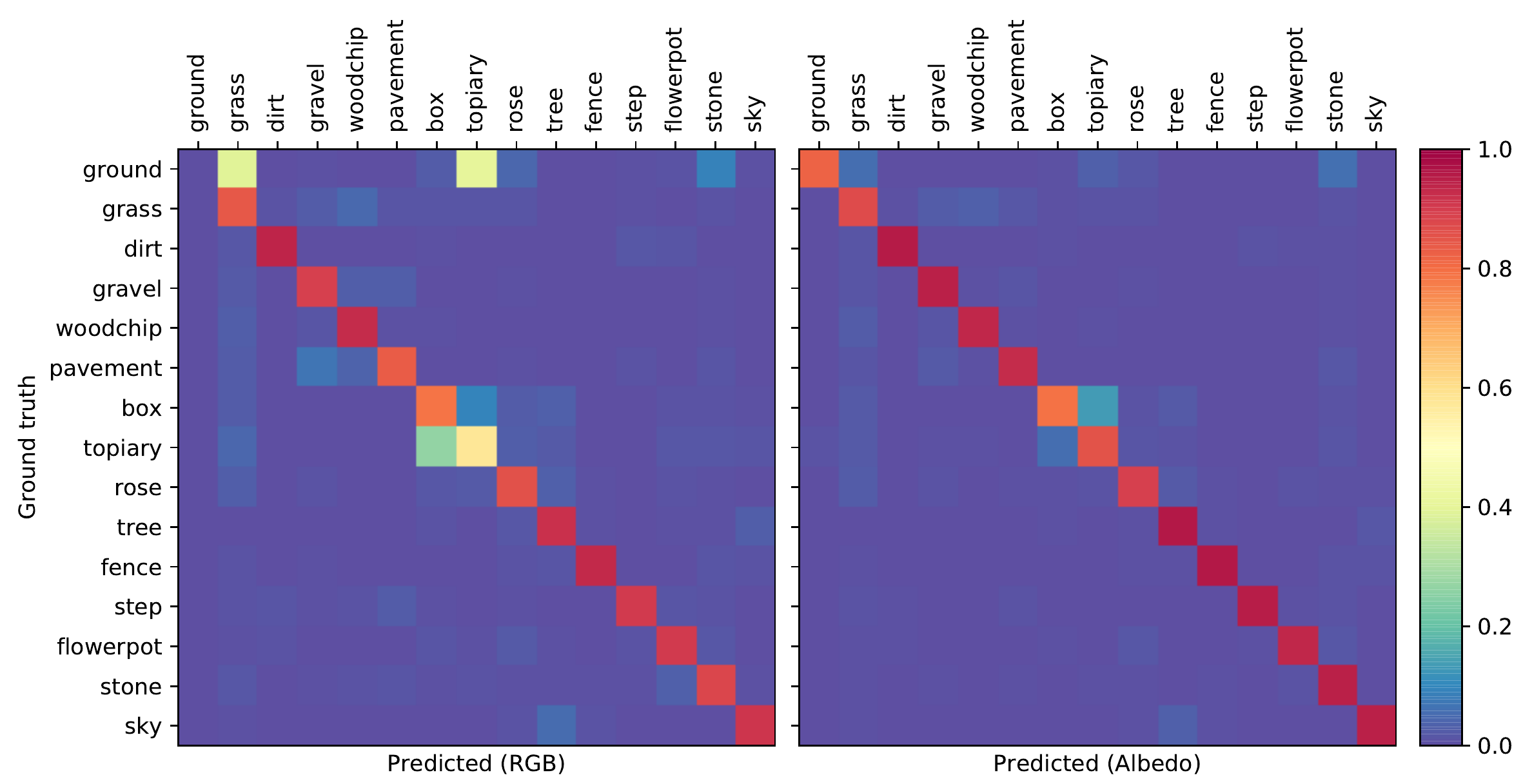}
    \caption{Confusion matrices for ($RGB-SegNet$) and ($Albedo-SegNet$)}
    \label{fig:cm1}
\end{figure}

In this experiment, we evaluate the performance of reflectance and RGB color images as input for semantic segmentation task. We train an off-the-shelf segmentation algorithm $SegNet$~\cite{segnet} using (i) ground-truth reflectance ($Albedo-SegNet$) and (ii) $RGB$ color images ($RGB-SegNet$); separately, and (iii) $RGB$ + reflectance ($Comb.-SegNet$); together, as input. The results are summarized in Table~\ref{Table:Segmentation} and illustrated in Figure~\ref{fig:segnet2}. Further, confusion matrices for ($RGB-SegNet$) and ($Albedo-SegNet$) are provided in Figure~\ref{fig:cm1}.

\begin{table}[]
\centering
\caption{Semantic segmentation accuracy using albedo and $RGB$ images as inputs. Using albedo images significantly outperforms $RGB$ images}
\label{Table:Segmentation}
\scalebox{0.8}{
\begin{tabular}{|c|c|c|c|c|l|l|l|l|l|l|l|l|l|l|l|l|}
\hline
Methodology & \multicolumn{1}{l|}{Global Pixel} & \multicolumn{1}{l|}{Class Average} & \multicolumn{1}{l|}{mIoU}
\\ \hline
$RGB-SegNet$ & 0.8743 & 0.6259 & 0.5217 
\\ \hline
$Comb.-SegNet$ & 0.8958 & 0.6607 & 0.5577 \\ \hline
$Albedo-SegNet$ & \textbf{0.9147} & \textbf{0.6739} & \textbf{0.5810}
\\ \hline
\end{tabular}
}
\end{table}

\noindent The results show that semantic segmentation algorithm highly benefits from illumination invariant intrinsic properties (i.e. reflectance). The combination ($Comb.-SegNet$) outperforms single RGB input ($RGB-SegNet$). On the other hand, the results with reflectance as single input ($Albedo-SegNet$) are superior to the results with inputs including $RGB$ color images in all metrics. The combined input ($Comb.-SegNet$) is not better than using only reflectance ($Albedo-SegNet$), because the network may be negatively influenced by the varying photometric cues introduced by the RGB input. Although the CNN framework may learn, to a certain degree, illumination invariance, it is not possible to cover all the variations caused by the illumination. Therefore, a full illumination invariant representation (i.e. reflectance) helps the CNN to improve semantic segmentation performance. Moreover, the confusion matrices show that the network is unable to distinguish a number of classes based on RGB input. Using reflectance, the same network gains the ability to correctly classify the ground class, as well as making fewer mistakes with similar-looking box and topiary classes.

\subsection{Influence of Semantic Segmentation on Intrinsic Decomposition}
In this experiment, we evaluate the performance of intrinsic image decomposition using ground-truth semantic segmentation labels as an extra source of information to the $RGB$ images. We compare the performance of intrinsic image decomposition trained with $RGB$ images ($RGB$) only as input and intrinsic decomposition trained with $RGB$ images and ground-truth semantic segmentation labels ($RGB+SegGT$) together as their input. As for $RGB+SegGT$, four input channels (i.e. $RGB$ color image and semantic segmentation labels) are provided as input. The results are summarized in Table~\ref{Table:InfluenceOfSeg}.

\begin{table}[]
\centering
\caption{The influence of semantic segmentation on intrinsic property prediction. Providing segmentation as an additional input ($RGB+SegGT$) clearly outperforms the approach of using only $RGB$ color images as their input}
\label{Table:InfluenceOfSeg}
\scalebox{0.65}{
\begin{tabular}{c|c|c|c|c|c|c|}
\cline{2-7}
 & \multicolumn{2}{c|}{MSE} & \multicolumn{2}{c|}{LMSE} & \multicolumn{2}{c|}{DSSIM} \\ \cline{2-7} 
& Alb & Shad & Alb & Shad & Alb & Shad \\ \hline
\multicolumn{1}{|c|}{$RGB$} & $0.0094~\pm~0.008$ & $0.0088~\pm~0.0078$ & $0.0679~\pm~0.0412$ & $0.0921~\pm~0.0582$ & $0.1310~\pm~0.0535$ & \textbf{0.1303$~\pm~$0.0495} \\ \hline
\multicolumn{1}{|c|}{$RGB+SegGT$} & \textbf{0.0076$~\pm~$0.0063} & \textbf{0.0078$~\pm~$0.0064} & \textbf{0.0620$~\pm~$0.0384} & \textbf{0.0901$~\pm~$0.0613} & \textbf{0.1141$~\pm~$0.0472} & $0.1312~\pm~0.0523$ \\ \hline
\end{tabular}
}
\end{table}

\begin{figure}[h]
    \centering
    \includegraphics[width=.9\textwidth]{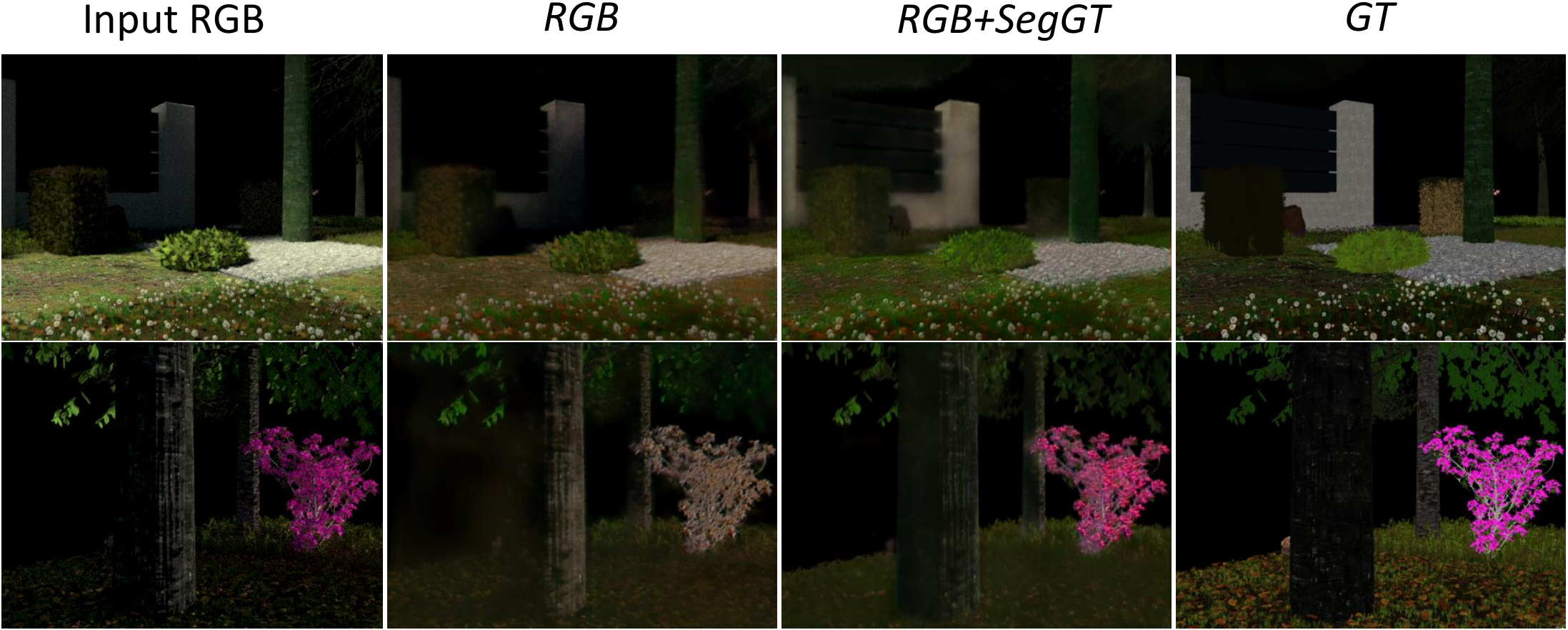}
    \caption{Columns 2 and 3 show that $RGB+SegGT$ is better in removing shadows and shading from the reflectance images, as well as preserving sharp object boundaries and vivid colors, and therefore is more similar to the ground truth}
    \label{fig:RGB+Seg}
\end{figure}

\noindent As shown in Table~\ref{Table:InfluenceOfSeg}, intrinsic image decomposition clearly benefits from segmentation labels. $RGB+SegGT$ outperforms $RGB$ in all metrics. DSSIM metric, accounting for the perceptual visual quality, shows the improvement on reflectance predictions, which indicates that the semantic segmentation process can act as an object boundary guidance map for reflectance prediction. A number of qualitative comparisons are shown for $RGB$ and $RGB+SegGT$ in Fig.~\ref{fig:RGB+Seg}.

\subsection{Joint Learning of Semantic Segmentation and Intrinsic Decomposition}
In this section, we evaluate the influence of joint learning on intrinsic image decomposition and semantic segmentation performances. We perform three experiments. First, we evaluate the effectiveness of joint learning of intrinsic properties and semantic segmentation considering semantic segmentation performance. Second, we evaluate the effectiveness of joint learning of intrinsic property and semantic segmentation to obtain intrinsic property prediction. Finally, we study the effects of the weights of the loss functions for the tasks.

\begin{figure}[t]
    \centering
    \includegraphics[width=.85\textwidth]{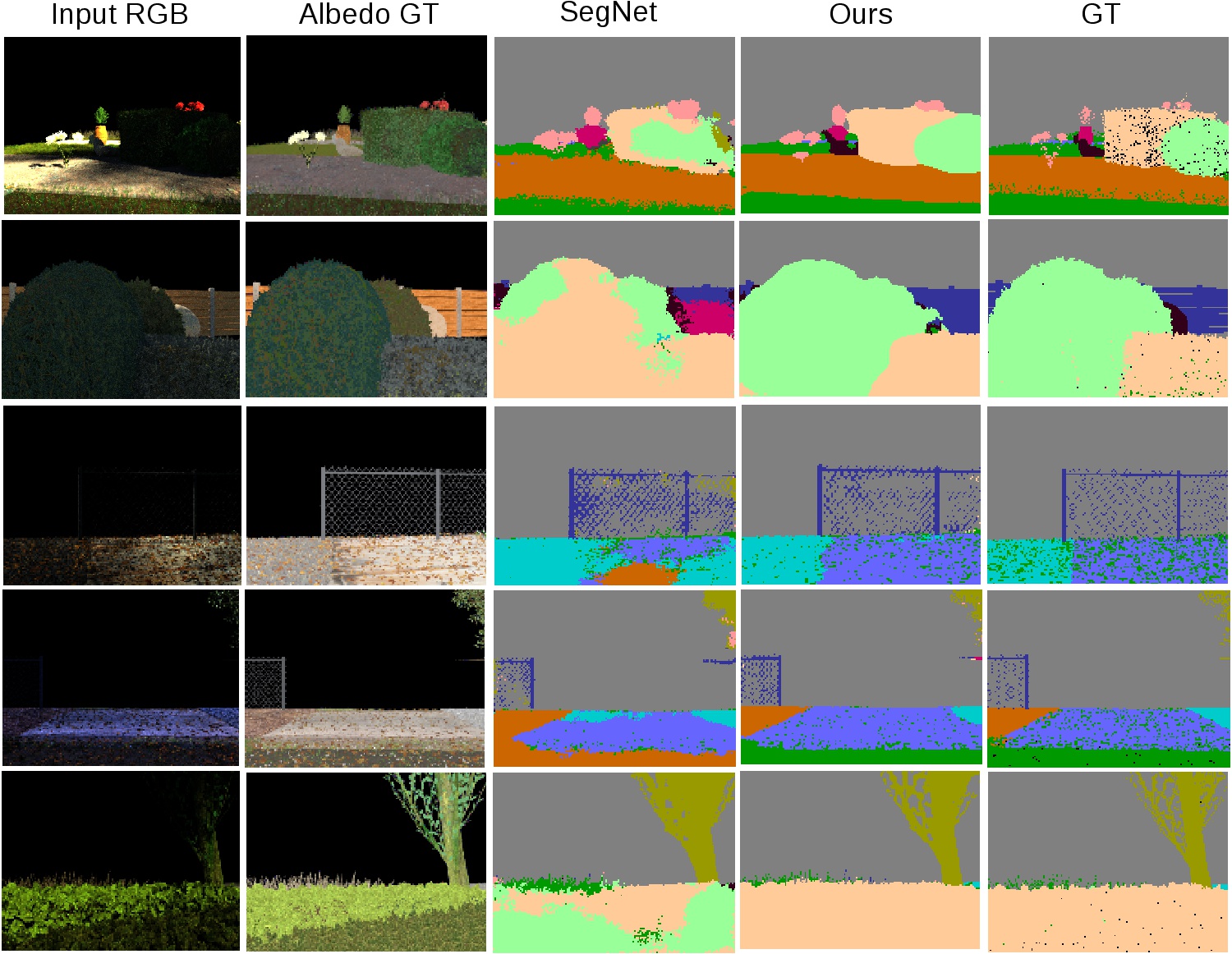}
    \caption{Proposed joint learning framework outperforms single task framework $SegNet$. Our method preserves the object shapes and boundaries better and is robust against varying lighting conditions}
    \label{fig:segmentation}
\end{figure}

\noindent \textbf{Experiment I.} In this experiment, we evaluate the performance of the proposed joint learning-based semantic segmentation algorithm ($Joint$), an off-the-shelf semantic segmentation algorithm~\cite{segnet} ($SegNet$) and the baseline of one encoder one decoder ShapeNet~\cite{shi} ($Single$). All CNNs receive $RGB$ color images as their input. $SegNet$ and $Single$ output only pixel level object class label predictions, whereas the proposed method predicts intrinsic property (i.e. reflectance and shading) in addition to the object class labels. We compare the accuracy of the models in Table~\ref{Table:JointSeg}. As shown in Table~\ref{Table:JointSeg}, the proposed joint learning framework outperforms the single task frameworks in all metrics. Further, visual comparison between $SegNet$ and the proposed $joint$ framework is provided in Fig.~\ref{fig:segmentation}. In addition, confusion matrices are provided in the supplementary material.

\begin{table}[]
\centering
\caption{Comparison of the semantic segmentation accuracy. The proposed joint learning framework outperforms the single task frameworks in all metrics}
\label{Table:JointSeg}
\scalebox{0.75}{
\begin{tabular}{|c|c|c|c|c|l|l|l|l|l|l|l|l|l|l|l|l|}
\hline
Methodology & \multicolumn{1}{l|}{Global Pixel} & \multicolumn{1}{l|}{Class Average} & mIoU
\\ \hline
$Single$ & 0.8022 & 0.4584 & 0.3659
\\ \hline
$SegNet$ & 0.8743 & 0.6259 & 0.5217
\\ \hline
$Joint$ & \textbf{0.9302} & \textbf{0.7055} & \textbf{0.6332}
\\ \hline
\end{tabular}
}
\end{table}

\noindent By analyzing the 3rd and 4th row of the figure, it can be derived that unusual lighting conditions negatively influence the results of the $SegNet$. In contrast, our proposed method is not effected by varying illumination due to the joint learning scheme. Furthermore, our method preserves object shapes and boundaries when compared to the $SegNet$ model (rows 1, 2 and 5). Note that the joint network does not perform any additional fine-tuning operations (e.g. CRF etc.). Additionally, $SegNet$ architecture is deeper than our proposed model. However, our method still outperforms $SegNet$. Finally, the joint network outperforms the single task cascade network; for mIoU 0.6332 vs. 0.5810, see Table~\ref{Table:Segmentation} and Table~\ref{Table:JointSeg}, as the joint scheme enforces to augment joint features.

\noindent \textbf{Experiment II.} In this experiment, we evaluate the performance of the proposed joint learning-based and the state-of-the-art intrinsic image decomposition algorithms~\cite{shi} ($ShapeNet$). Both CNNs receive $RGB$ color images as input. $ShapeNet$ outputs only intrinsic properties (i.e. reflectance and shading), whereas the proposed method predicts pixel level object class labels as well as intrinsic properties. We train $ShapeNet$ and the proposed method using ground-truth reflectance and shading labels on the training set of the proposed dataset. We compare the accuracy of $ShapeNet$ and the proposed method in Table~\ref{Table:JointInt}. 

\begin{table}[h]
\centering
\caption{Influence of joint learning on intrinsic property prediction}
\label{Table:JointInt}
\scalebox{0.65}{
\begin{tabular}{c|c|c|c|c|c|c|c|c|c|c|c|c|}
\cline{2-7}
 & \multicolumn{2}{c|}{MSE} & \multicolumn{2}{c|}{LMSE} & \multicolumn{2}{c|}{DSSIM}  \\ \cline{2-7} 
& Alb & Shad & Alb & Shad & Alb & Shad \\ \hline
\multicolumn{1}{|c|}{$ShapeNet$} & $0.0094~\pm~0.0080$ & $0.0088~\pm~0.0078$ & $0.0679~\pm~0.0412$ & $0.0921~\pm~0.0582$ & $0.1310~\pm~0.0535$ & $0.1303~\pm~0.0495$ \\ \hline
\multicolumn{1}{|c|}{Int.-Seg. Joint} & $\textbf{0.0030}~\pm~\textbf{0.0040}$ & $\textbf{0.0030}~\pm~\textbf{0.0024}$ & $\textbf{0.0373}~\pm~\textbf{0.0356}$ & $\textbf{0.0509}~\pm~\textbf{0.0395}$ & $\textbf{0.0753}~\pm~\textbf{0.0399}$ & $\textbf{0.0830}~\pm~\textbf{0.0381}$  \\ \hline
\end{tabular}
}
\end{table}

\begin{figure}[h!]
    \centering
    \includegraphics[width=.9\textwidth]{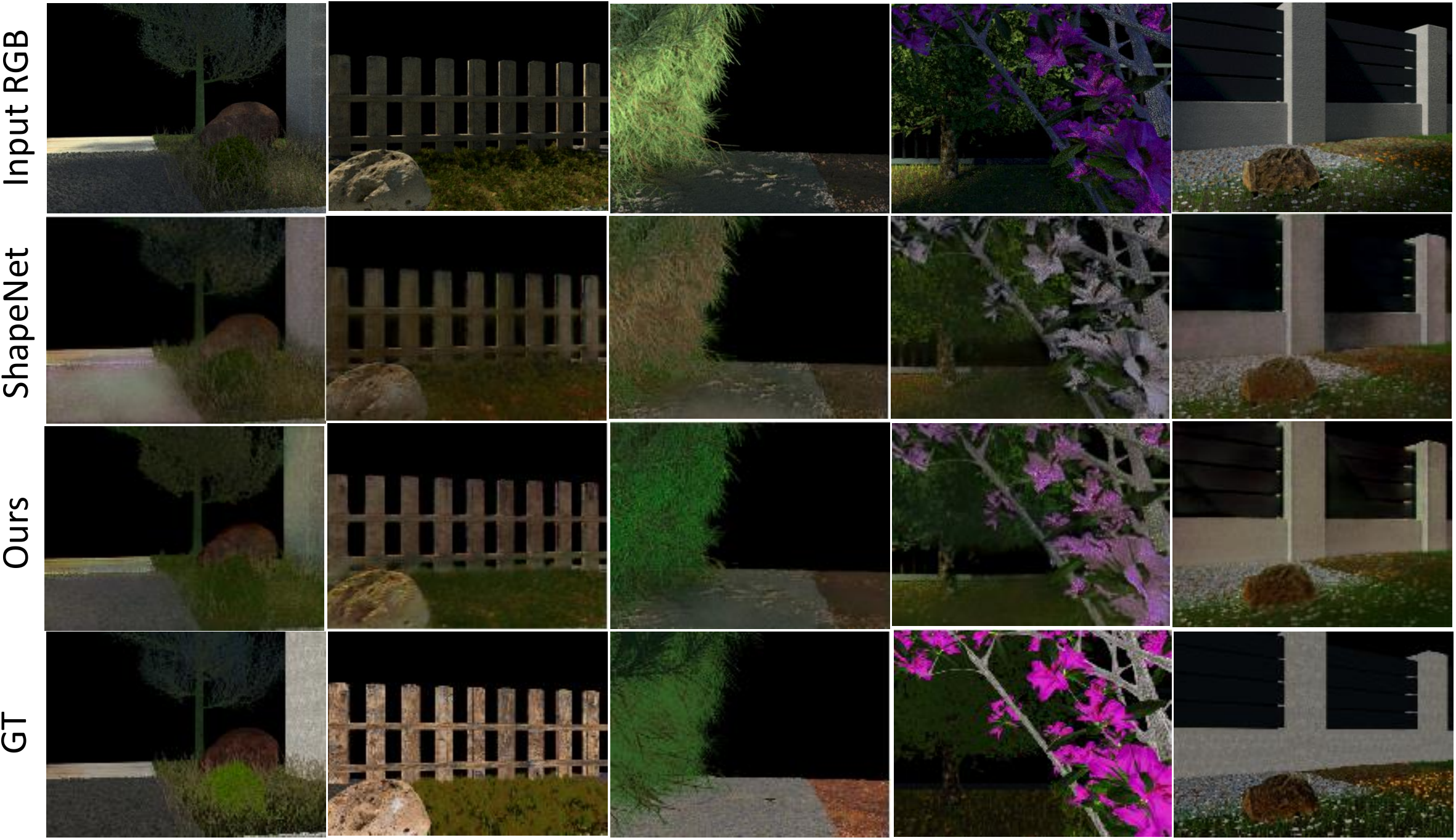}
    \caption{The first two columns illustrate that the proposed method provides sharper outputs especially at object boundaries than $ShapeNet$. The 3rd and 4th columns show that the proposed method predicts colours that are closer to the ground truth reflectance. The last column shows that the proposed method handles sharp cast shadows better than $ShapeNet$}
    \label{fig:JointInt}
\end{figure}

\noindent As shown in Table~\ref{Table:JointInt}, the performance of the proposed joint learning framework outperforms single task learning ($ShapeNet$) in all the metrics for reflectance (albedo) and shading estimation. Further, our joint model obtains lower standard deviation values. To give more insight on reflectance prediction performances, a number of visual comparisons between $ShapeNet$ and the proposed $joint$ framework are given in Fig.~\ref{fig:JointInt}. In the figure, (the first two columns) it can be derived that the semantic segmentation process acts as an object boundary guidance map for the intrinsic image decomposition task by enhancing cues to differentiate between reflectance and occlusion edges in a scene. Hence, object boundaries are better preserved by the proposed method (e.g. the separation between pavement and ground in the first image and the space between fences in the second image). In addition, information about an object reveals strong priors about it's intrinsic properties. Each object label adopts to a constrained color distribution. That can be observed in third and fourth columns. Semantic segmentation guides intrinsic image decomposition process by yielding the trees to be closer to green and flowers to be closer to pink. Moreover, for class-level intrinsics, the best improvement (3.3 times better) is obtained by \emph{concrete step blocks}, which have achromatic colors. Finally, as in segmentation, the joint network outperforms the single task cascade network, see Table~\ref{Table:InfluenceOfSeg} and Table~\ref{Table:JointInt}.

\noindent \textbf{Experiment III.} In this experiment, we study the effects of the weightings of the loss functions. As the cross entropy loss is an order of magnitude higher than the SMSE loss, we first normalize them by multiplying the intrinsic loss by 100. Then, we evaluate different weights on top of the normalization ($SMSE\times100\times w$). See Table~\ref{Table:weights} for the results. If higher weights are assigned to intrinsics, they both jointly increase. However, weights which are too high, negatively influence the mIoU values. Therefore, $w = 2$ appears to be the proper setting for both tasks.

\begin{table}[]
\centering
\caption{Influence of the weighting of the loss functions. SMSE loss is weighted by ($SMSE\times100\times w$). $w = 2$ appears to be the proper setting for both tasks} 
\label{Table:weights}
\scalebox{0.68}{
\begin{tabular}{|c|c|c|c|c|c|c|c|c|}
\hline
\multirow{2}{*}{$\omega$} & \multicolumn{2}{c|}{Segmentation} & \multicolumn{2}{c|}{MSE} & \multicolumn{2}{c|}{LMSE} & \multicolumn{2}{c|}{DSSIM} \\ \cline{2-9} 
 & Global & mIoU & Alb & Shad & Alb & Shad & Alb & Shad \\ \hline
$0.01$ & $0.9179$ & $0.567$ & $0.0083~\pm~0.0068$ & $0.0083~\pm~0.0072$ & $0.0650~\pm~0.0412$ & $0.0920~\pm~0.0611$ & $0.1224~\pm~0.0498$ & $0.1343~\pm~0.0545$ \\ \hline
$0.5$ & $0.7038$ & $0.512$ & $0.0038~\pm~0.0037$ & $0.0035~\pm~0.0027$ & $0.0398~\pm~0.0311$ & $0.0550~\pm~0.0416$ & $0.1633~\pm~0.0538$ & $0.1353~\pm~0.0497$ \\ \hline
$1$ & $0.9048$ & $0.533$ & $0.0044~\pm~0.0041$ & $0.0044~\pm~0.0036$ & $0.0477~\pm~0.0352$ & $0.0655~\pm~0.0474$ & $0.0926~\pm~0.0445$ & $0.1040~\pm~0.0421$ \\ \hline
$2$ & $0.9302$ & $0.633$ & $0.0030~\pm~0.0040$ & $0.0030~\pm~0.0024$ & $0.0373~\pm~0.0356$ & $0.0509~\pm~0.0395$ & $0.0753~\pm~0.0399$ & $0.0830~\pm~0.0381$ \\ \hline
$4$ & $0.9334$ & $0.611$ & $0.0028~\pm~0.3300$ & $0.0028~\pm~0.0023$ & $0.0356~\pm~0.02997$ & $0.0491~\pm~0.04081$ & $0.0716~\pm~0.03804$ & $0.0695~\pm~0.0357$ \\ \hline
\end{tabular}
}
\end{table}

\subsection{Real World Outdoor Dataset}
\begin{figure}[t]
    \centering
    \includegraphics[width=.9\textwidth]{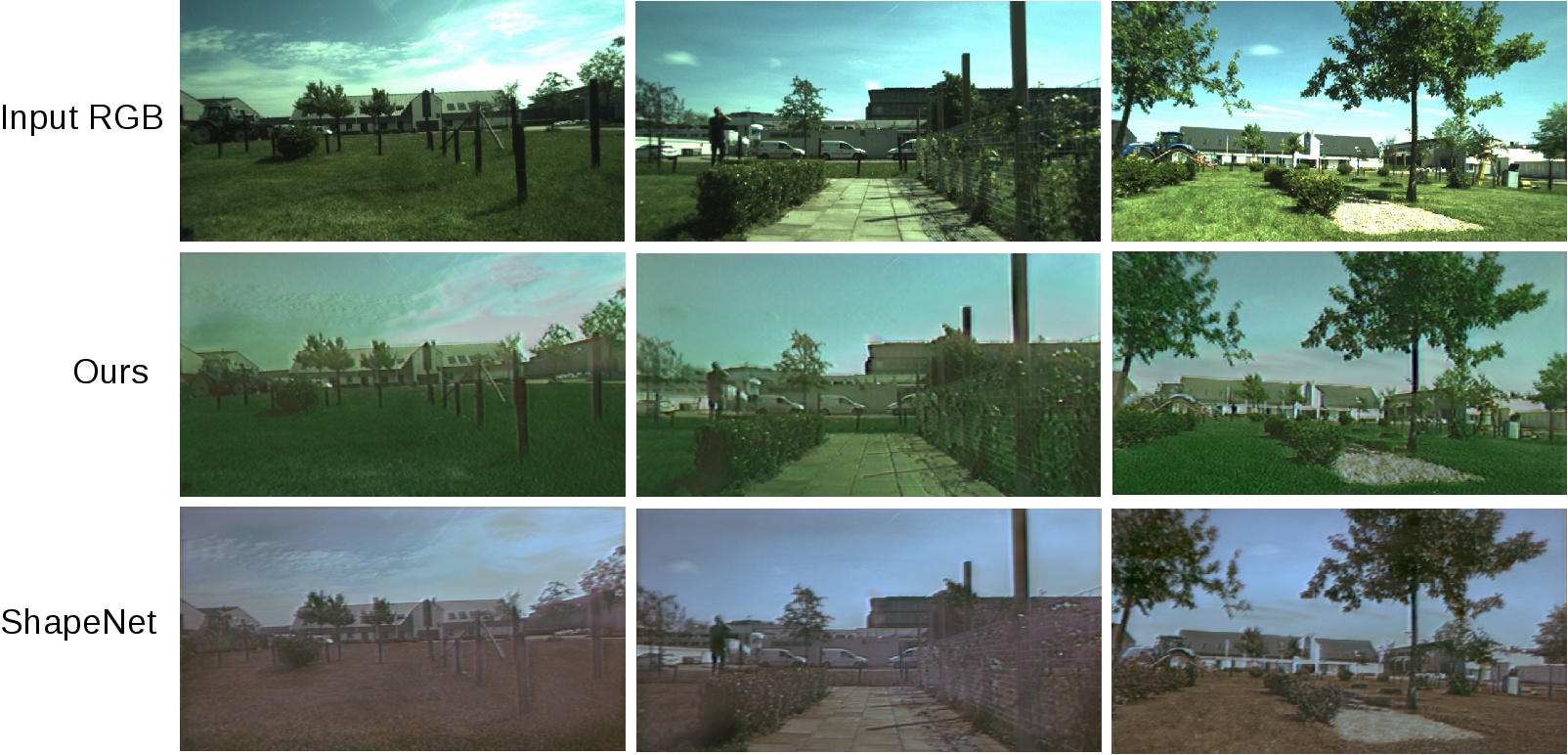}
    \caption{Evaluation on real world garden images. We observe that our proposed method capture better colors and sharper outputs compared with \cite{shi}}
    \label{fig:real}
\end{figure}

Finally, our model is evaluated on real world garden images provided by the {\em 3D Reconstruction meets Semantics challenge} \cite{rms_challange}. The images are captured by a robot driving through a semantically-rich garden with fine geometric details. Results of \cite{shi} are provided as a visual comparison on the performance in Fig.~\ref{fig:real}. It shows that our method generates better results on real images with sharper reflectance images having more vivid and realistic colors. Moreover, our method mitigates sharp shadow effects better. Note that our model is trained fully on synthetic images and still provides satisfactory results on real, natural scenes. For semantic segmentation comparison, we fine-tuned SegNet~\cite{segnet} and our approach on the real world dataset after pre-training on the garden dataset. Since we only have the ground-truth for segmentation, we (only) unfreeze the segmentation branch. Results show that SegNet and our approach obtain 0.54 and 0.54 for mIoU and a global pixel accuracy of 0.85 and 0.88 respectively. Note that our model is much smaller in size and predicts the intrinsics together with the segmentation. More results are provided in the supplementary material.

\section{Conclusion}
Our approach jointly learns intrinsic image decomposition and semantic segmentation. New CNN architectures are proposed for joint learning, and single intrinsic-for-segmentation and segmentation-for-intrinsic learning. A dataset of 35K synthetic images of natural environments has been created with corresponding albedo and shading (intrinsics), and semantic labels (segmentation). The experiments show joint performance benefit when performing the two tasks (intrinsics and semantics) in joint manner for natural scenes.

\noindent \textbf{Acknowledgements:} This project was funded by the EU Horizon 2020 program No. 688007 (TrimBot2020). We thank Gjorgji Strezoski for his contributions to the website.

\clearpage

\bibliographystyle{splncs}
\bibliography{egbib}

\newpage

\begin{center}
\textbf{\large Joint Learning of Intrinsic Images and Semantic Segmentation Supplementary Material}
\end{center}

\setcounter{equation}{0}
\setcounter{figure}{0}
\setcounter{table}{0}
\setcounter{section}{0}
\setcounter{page}{1}
\makeatletter
\renewcommand{\theequation}{S\arabic{equation}}
\renewcommand{\thefigure}{S\arabic{figure}}
\renewcommand{\thesection}{S\arabic{section}}

\section{Implementation Details}
Our models are implemented using the Adadelta~\citeS{supp_adadelta} optimizer with learning rate of 0.01. Convolution weights are initialized using a normal distribution with a weight decay factor of 1e-9. The input to the networks are fixed to a resolution of 352 $\times$ 480. The images are normalized to the range of [0, 1] as the pre-processing step. The batch sizes are fixed at 16 for all experiments. In addition, in the last decoder block, the feature dimension is reduced to the expected output dimensions. For the semantic segmentation task, the loss is weighted per class, since the classes are not equally distributed. Furthermore, for the output of the semantic segmentation task, the feature dimensions are unchanged and is simply convolved at the same feature dimension one additional time. This produces an output corresponding to the 16 class labels. 
\newline
\noindent \textbf{Baseline network architecture:} The encoder part is composed of 6 convolution blocks with 3 $\times$ 3 kernels and stride of 2 having [16, 32, 64, 128, 256, 256] feature maps. The encoder part is mirrored to build the decoder. Before convolving the feature maps of the decoder part, previous layer's feature maps are first up-sampled and concatenated with their corresponding encoder features. All convolutions are followed by batch normalization~\citeS{supp_batchnorm} and ReLU.
\newline
\noindent \textbf{Gamma parameters for SMSE:} For all the experiments involving intrinsic image decomposition task, to form the combined MSE and SMSE loss, we followed the setup of~\citeS{supp_shi} and set $\gamma_{SMSE}$ to 0.95 and $\gamma_{MSE}$ to 0.05 for Equation 6 in the main manuscript. Nonetheless, we conducted a small experiment to see the effect of the gamma parameters for SMSE. Table~\ref{tab:gamma} provides the average intrinsic image decomposition errors. The small experiment suggest that giving higher weight to $\gamma_{SMSE}$ tends to improve the results.

\begin{table}[h]
  \centering
  \scalebox{1}{
  \begin{tabular}{ c||c||c||c|}
    \multirow{2}{*}{ } &
      \multicolumn{1}{c||}{MSE}  &
      \multicolumn{1}{c||}{LMSE} &
      \multicolumn{1}{c|}{DSSIM} \\ \hline
       $\gamma_{SMSE}$ = 0.95 & 0.0083 & 0.0785 & 0.1284 \\
       $\gamma_{SMSE}$ = 0.99 & \textbf{0.0043} & \textbf{0.0631} & \textbf{0.1068} \\
  \end{tabular}}
  \newline
  \caption {Effect of gamma parameters for SMSE}
  \label{tab:gamma}
\end{table}

\newpage

\section{Confusion Matrix for Joint Learning of Semantic Segmentation and Intrinsic Decomposition}
\begin{figure}[h]
    \centering
    \includegraphics[width=0.8\textwidth]{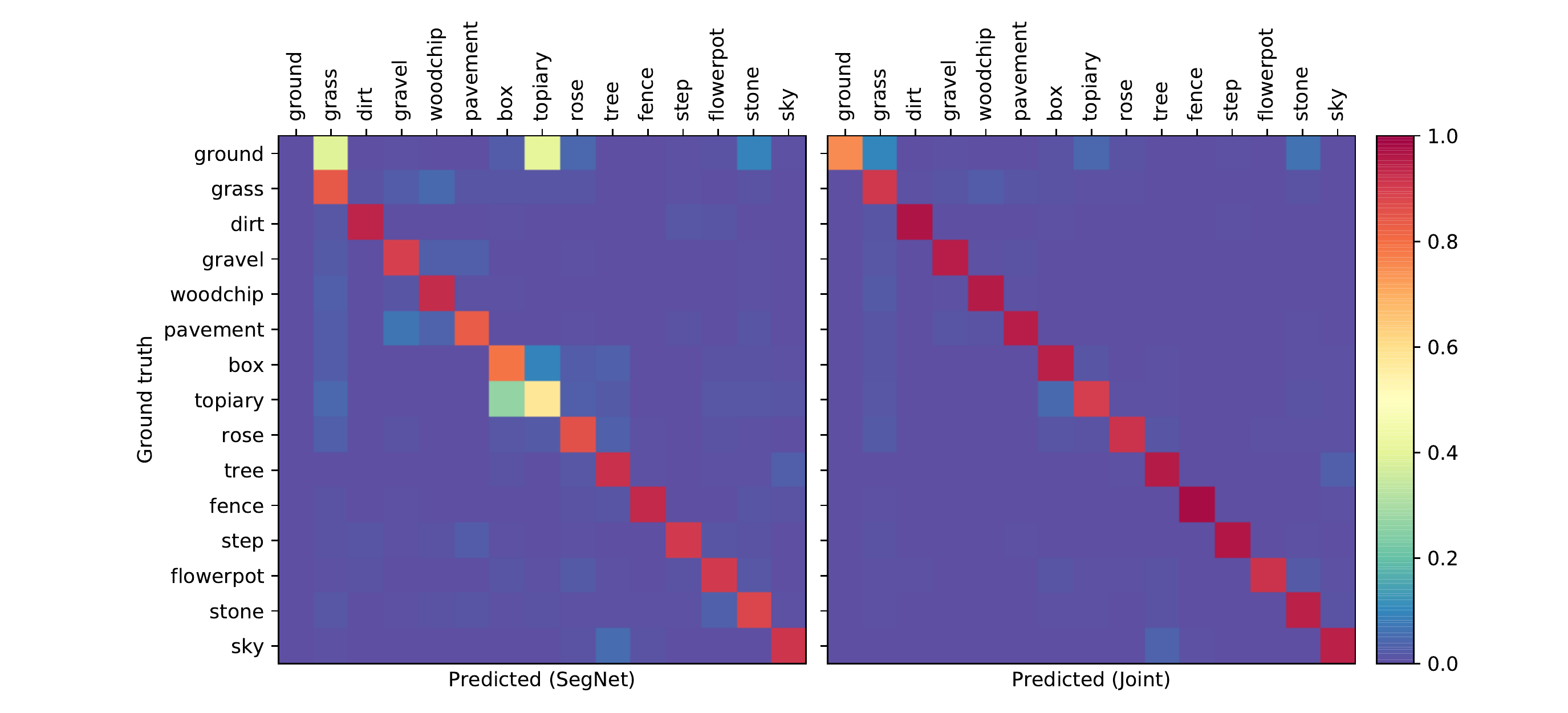}
    \caption{Confusion matrices for $SegNet$ and proposed $Joint$ model. Results suggest that close-color classes under different lighting conditions is further improved by joint learning and most of the small confusions are eliminated}
    \label{fig:cm2}
\end{figure}

Confusion matrices for $SegNet$ and proposed $Joint$ model are provided in Figure~\ref{fig:cm2}. Confusion matrices show that the ability to distinguish close-color classes under different lighting conditions is further improved by joint learning. Similar to the case with using albedo as input for SegNet architecture, joint learning also improves the semantic segmentation performance significantly with certain classes. For the ground class, confusion is reduced remarkably by also learning intrinsics. Likewise, similar looking (in terms of shape and color) box and topiary classes are also better distinguished. In addition, most of the small confusions are eliminated.

\section{Results in Higher Resolutions}
In this part of the supplementary material, the results of reflectance prediction in higher resolution are presented for better visual comparisons.

\subsection{Influence of Semantic Segmentation on Intrinsic Image Decomposition}
\begin{figure}[h]
    \centering
    \includegraphics[width=.9\textwidth]{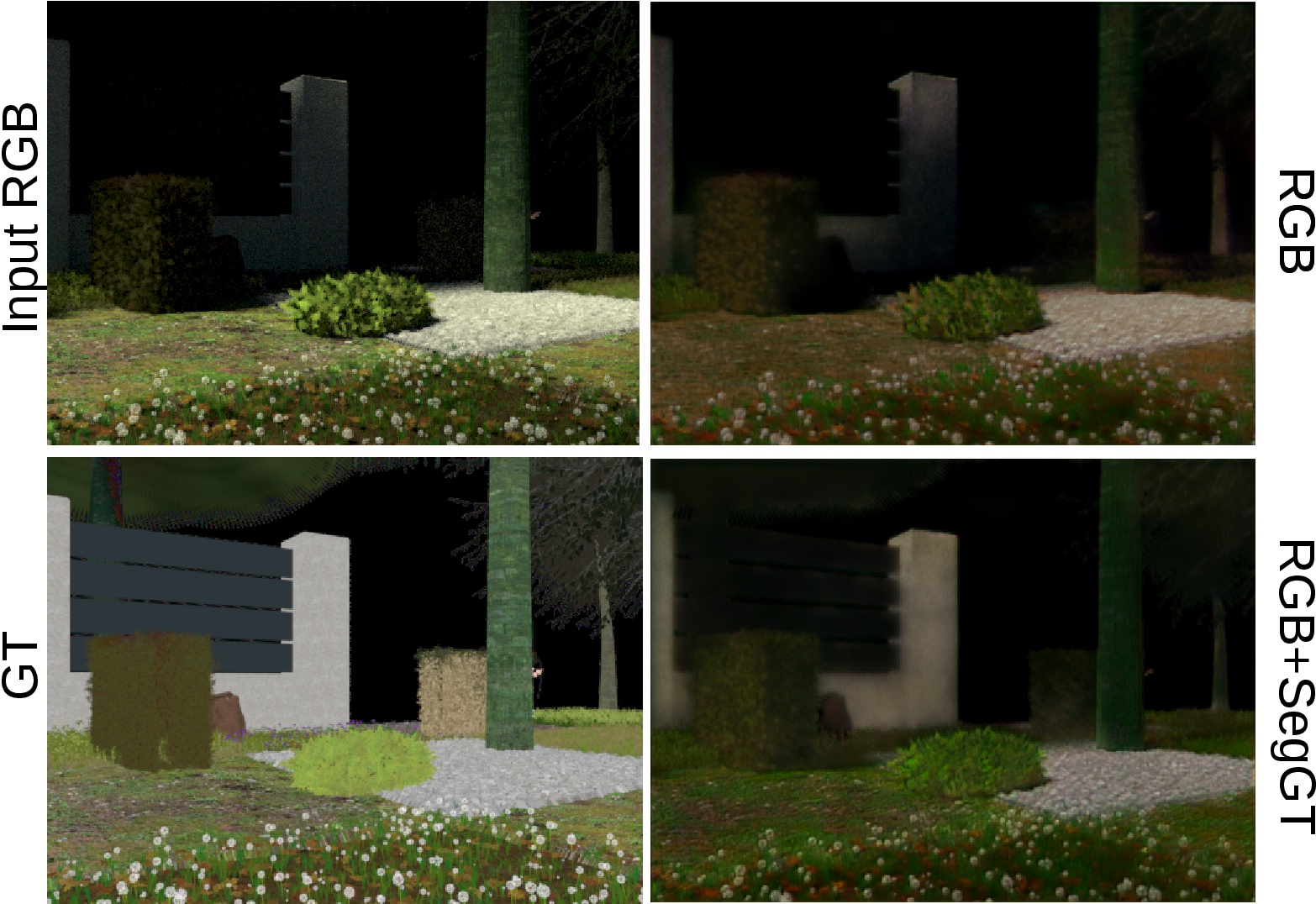}
    \caption{Higher resolution images on the influence of semantic segmentation on intrinsic image decomposition (1)}
    \label{fig:11}
\end{figure}

\begin{figure}[h]
    \centering
    \includegraphics[width=.9\textwidth]{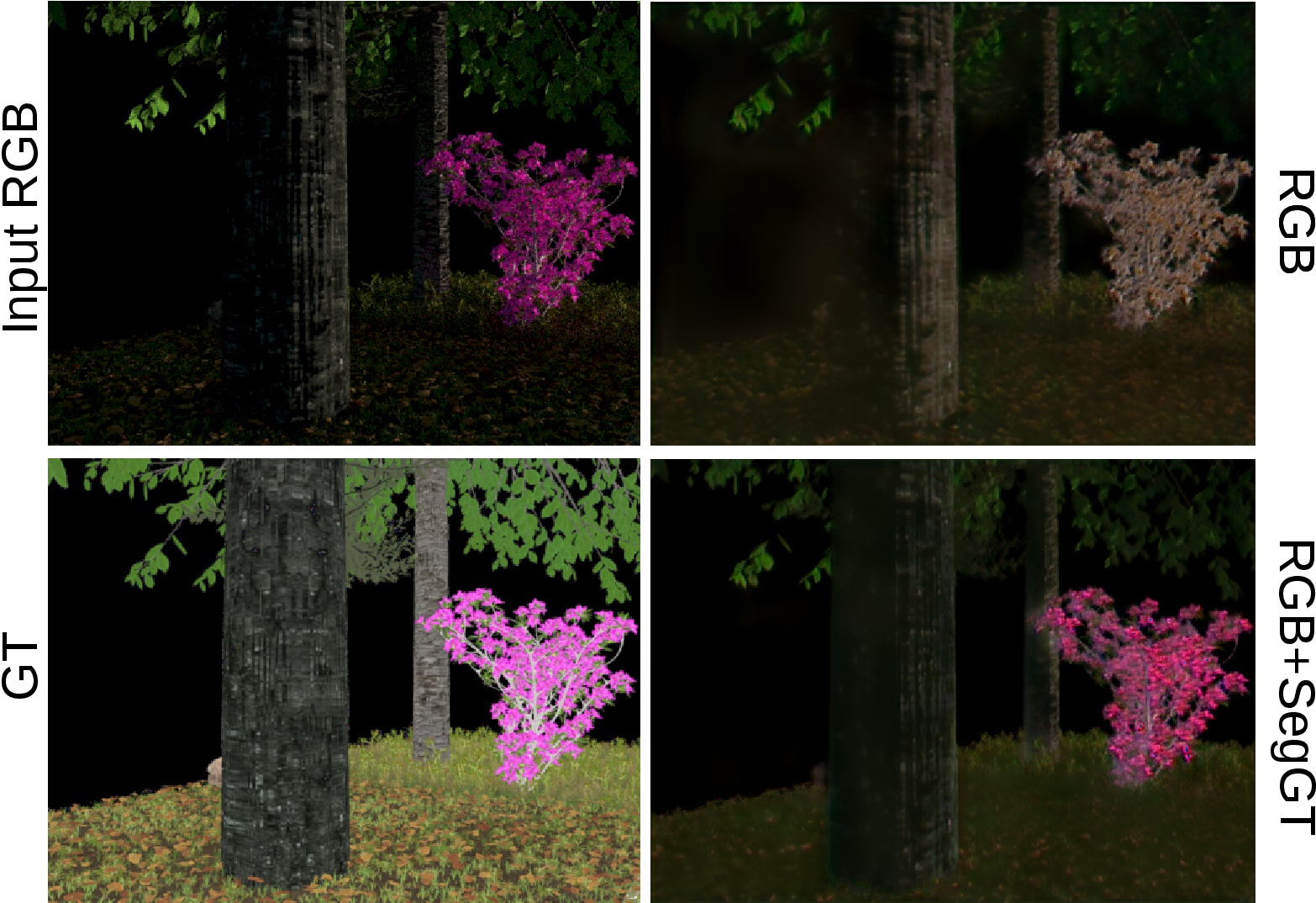}
    \caption{Higher resolution images on the influence of semantic segmentation on intrinsic image decomposition (2)}
    \label{fig:12}
\end{figure}

In this experiment, we evaluate the performance of intrinsic image decomposition using ground-truth semantic segmentation labels as an extra source of information to the $RGB$ color images. Qualitative comparison between predictions made from $RGB$ images as input ($RGB$), against predictions made from $RGB$ along with segmentation labels as input ($RGB+SegGT$) are provided in higher resolutions in Fig.~\ref{fig:11} and Fig.~\ref{fig:12}.

\subsection{Joint Learning of Semantic Segmentation and Intrinsic Decomposition}
\begin{figure}[h]
    \centering
    \includegraphics[width=.9\textwidth]{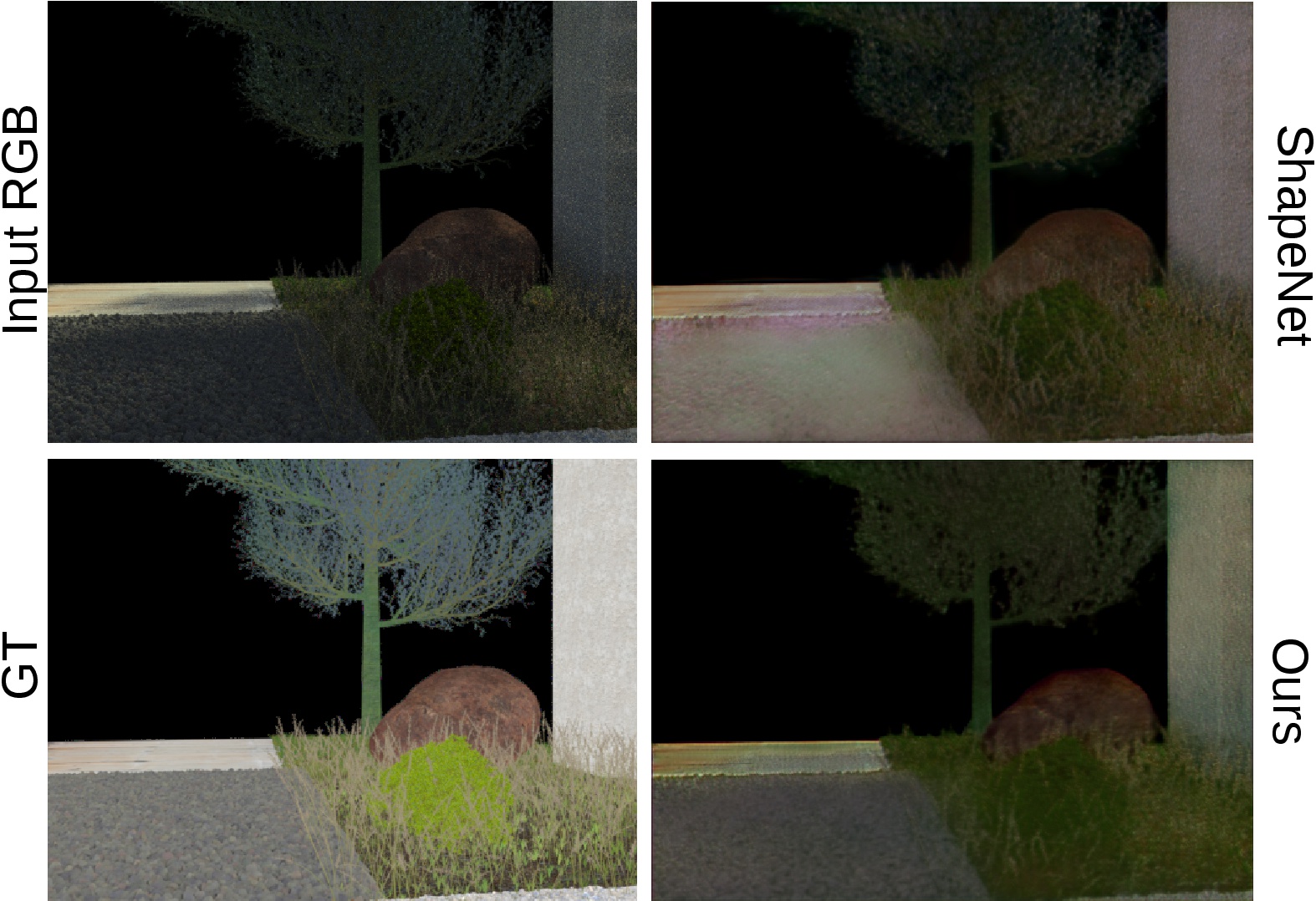}
    \caption{Higher resolution images on the comparison of the proposed method with $ShapeNet$ (1)}
    \label{fig:21}
\end{figure}

\begin{figure}[h]
    \centering
    \includegraphics[width=.9\textwidth]{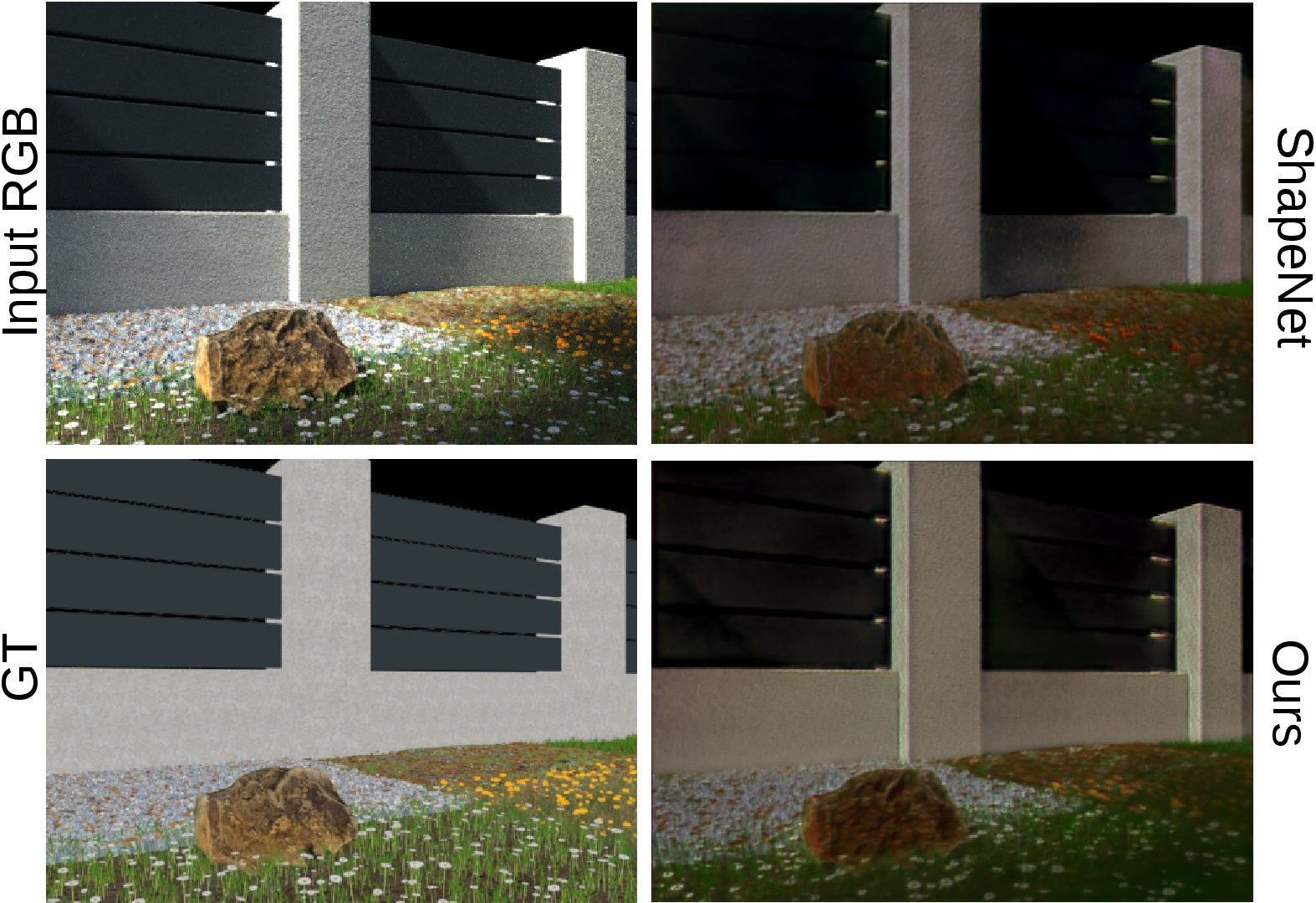}
    \caption{Higher resolution images on the comparison of the proposed method with $ShapeNet$ (2)}
    \label{fig:22}
\end{figure}

\begin{figure}[h]
    \centering
    \includegraphics[width=.9\textwidth]{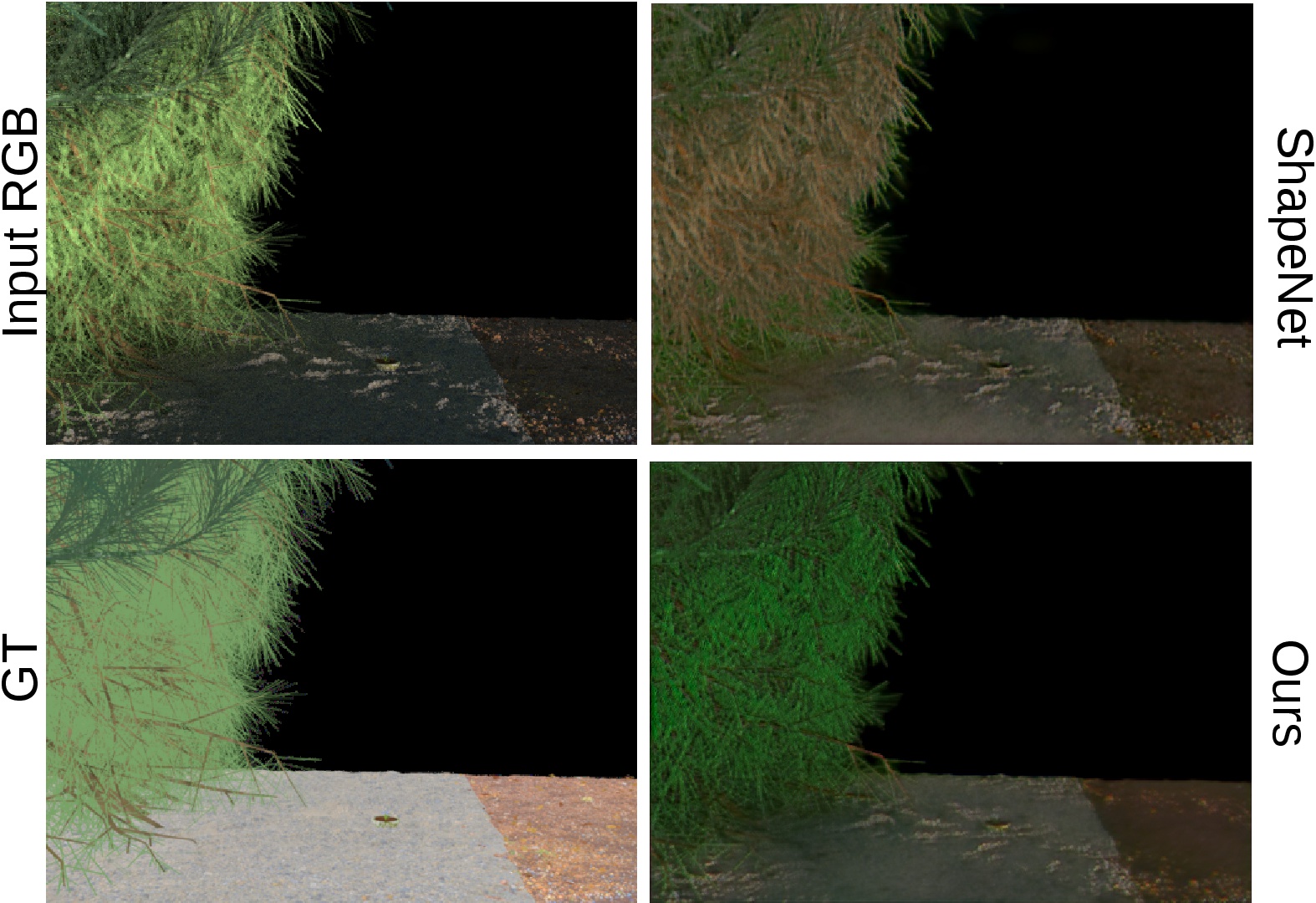}
    \caption{Higher resolution images on the comparison of the proposed method with $ShapeNet$ (3)}
    \label{fig:23}
\end{figure}

\begin{figure}[h]
    \centering
    \includegraphics[width=.9\textwidth]{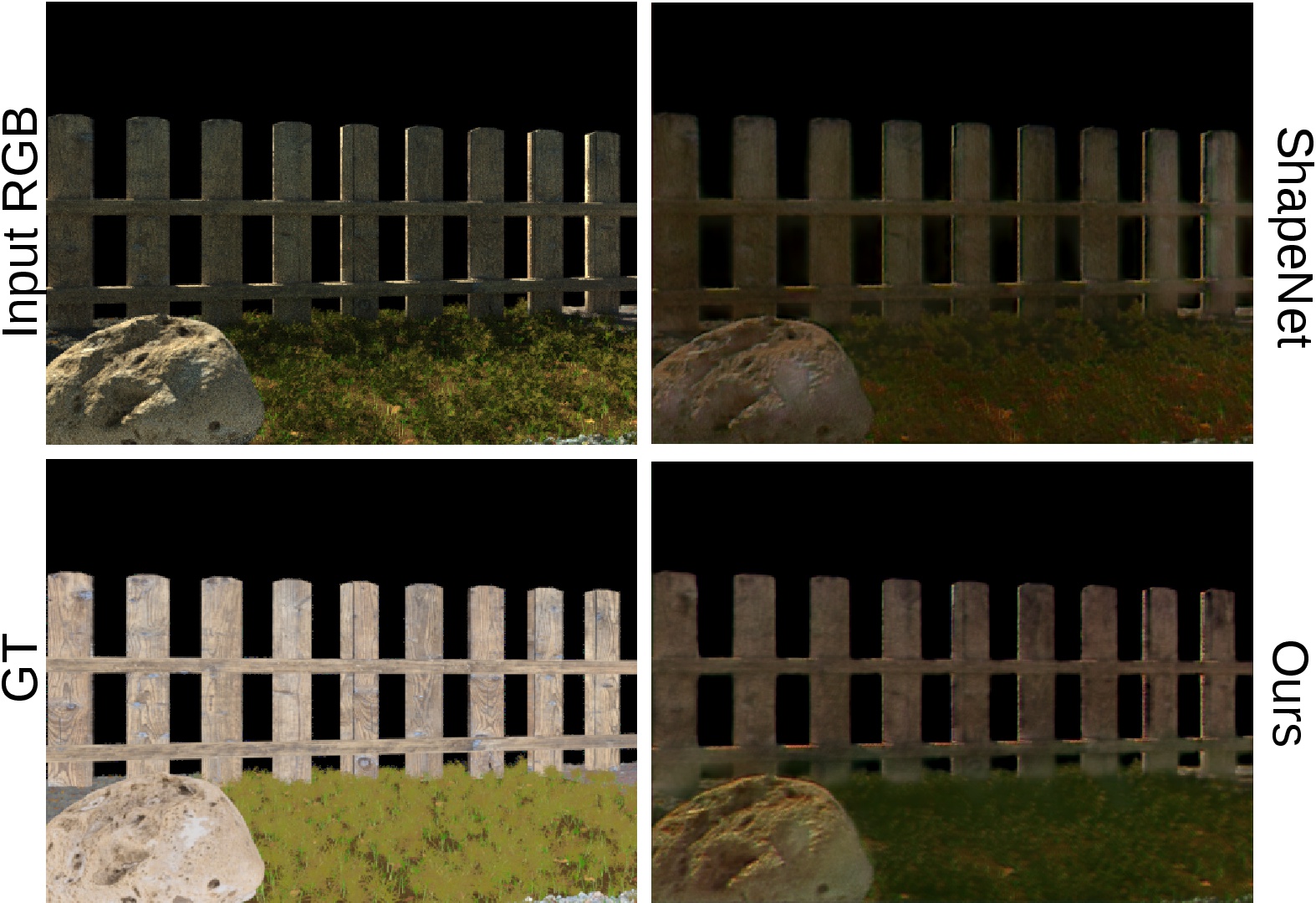}
    \caption{Higher resolution images on the comparison of the proposed method with $ShapeNet$ (4)}
    \label{fig:24}
\end{figure}

\begin{figure}[h]
    \centering
    \includegraphics[width=.9\textwidth]{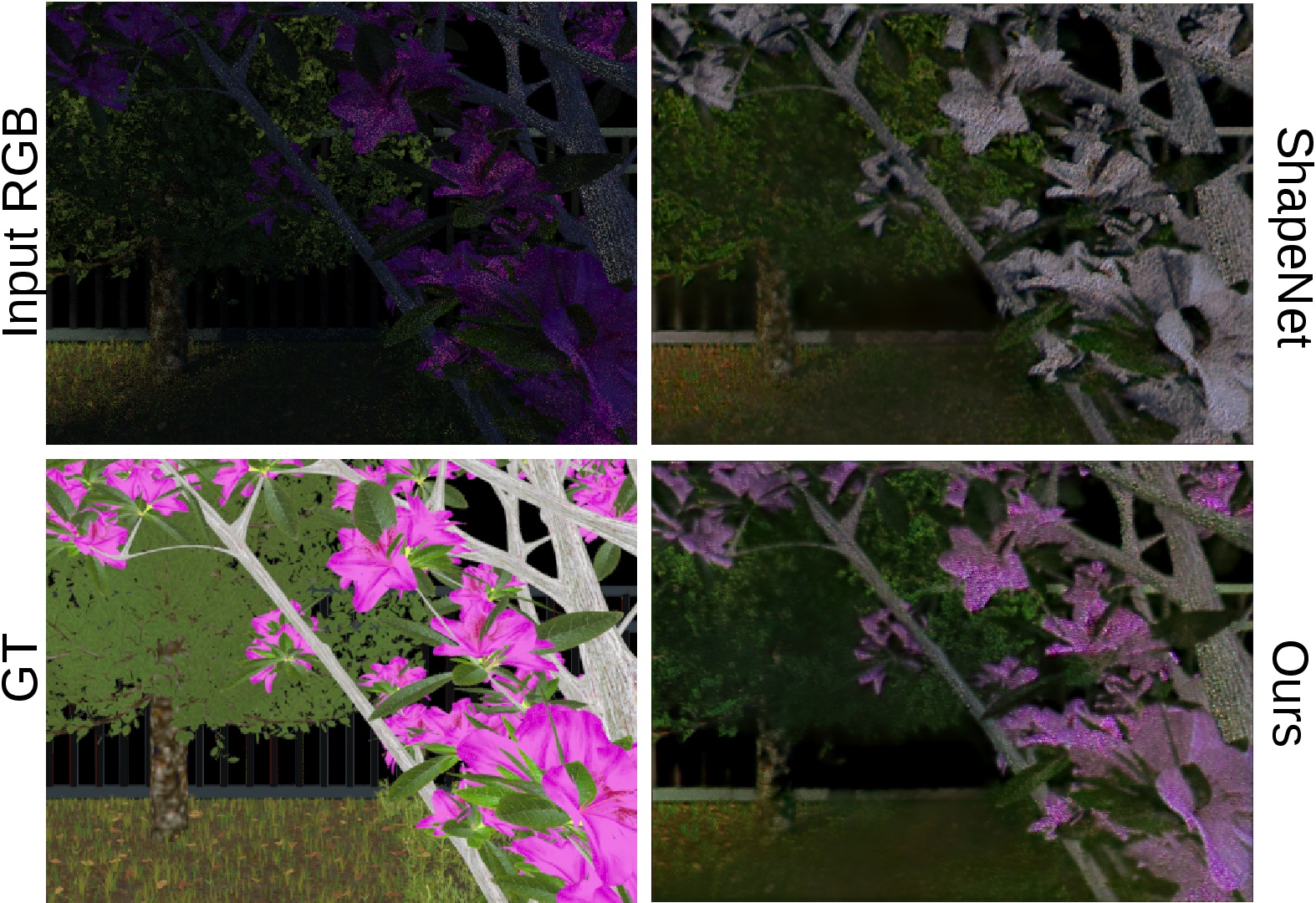}
    \caption{Higher resolution images on the comparison of the proposed method with $ShapeNet$ (5)}
    \label{fig:25}
\end{figure}

In this experiment, we evaluate the performance of the proposed joint learning-based and the a state-of-the-art intrinsic image decomposition algorithm $ShapeNet$~\citeS{supp_shi}. Both CNNs receive $RGB$ color images as input. The proposed method provides sharper outputs especially at object boundaries and predicts colours that are closer to the ground truth reflectance. Higher resolution results are provided from Fig.~\ref{fig:21} to Fig.~\ref{fig:25}.

\section{More Results on Real World Images}
\begin{figure}[h]
    \centering
    \includegraphics[width=.9\textwidth]{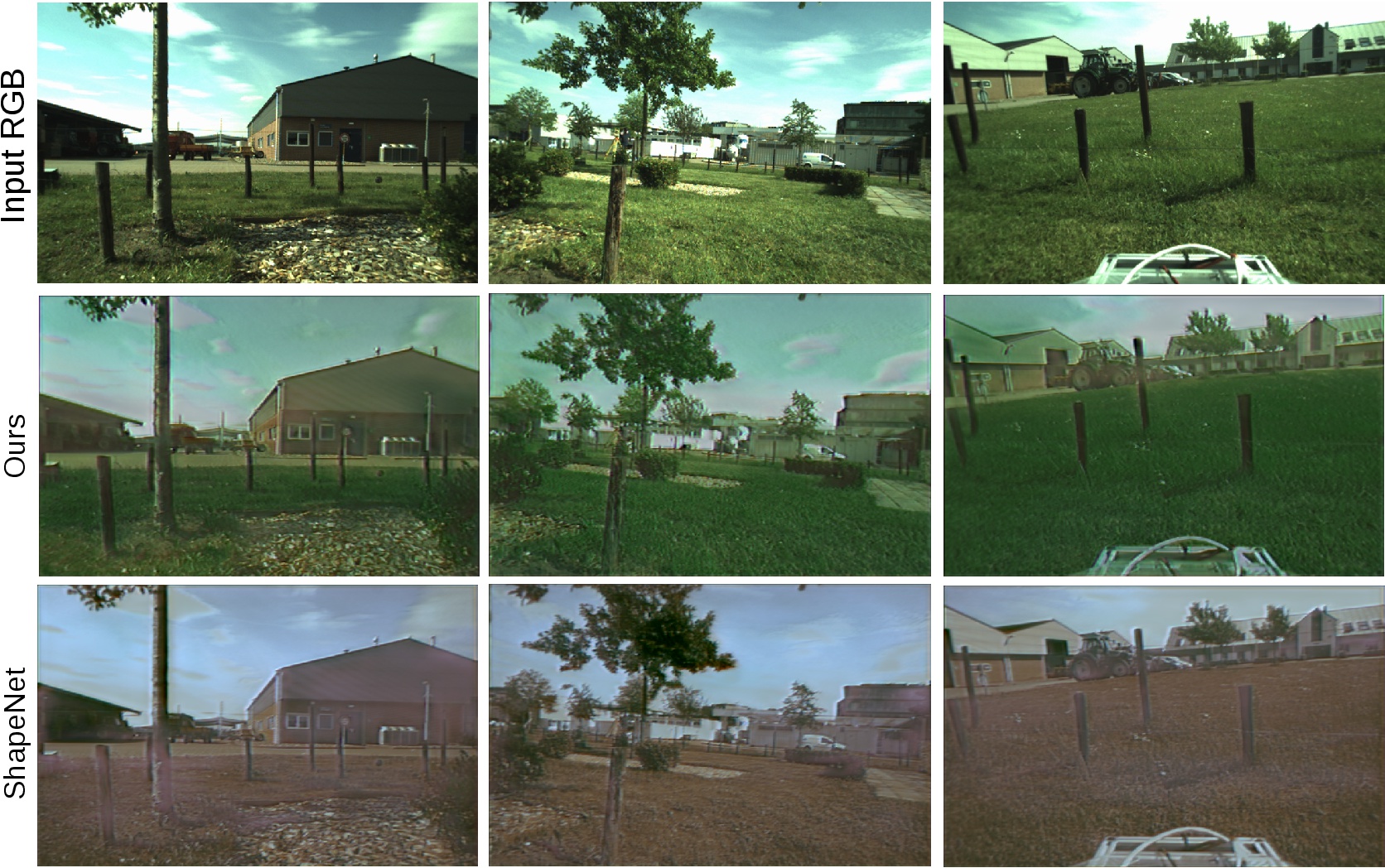}
    \caption{Intrinsic image decomposition evaluation on real world garden images. We observe that our proposed method captures better colors and sharper outputs}
    \label{fig:3}
\end{figure}

\begin{figure}[h]
    \centering
    \includegraphics[width=.9\textwidth]{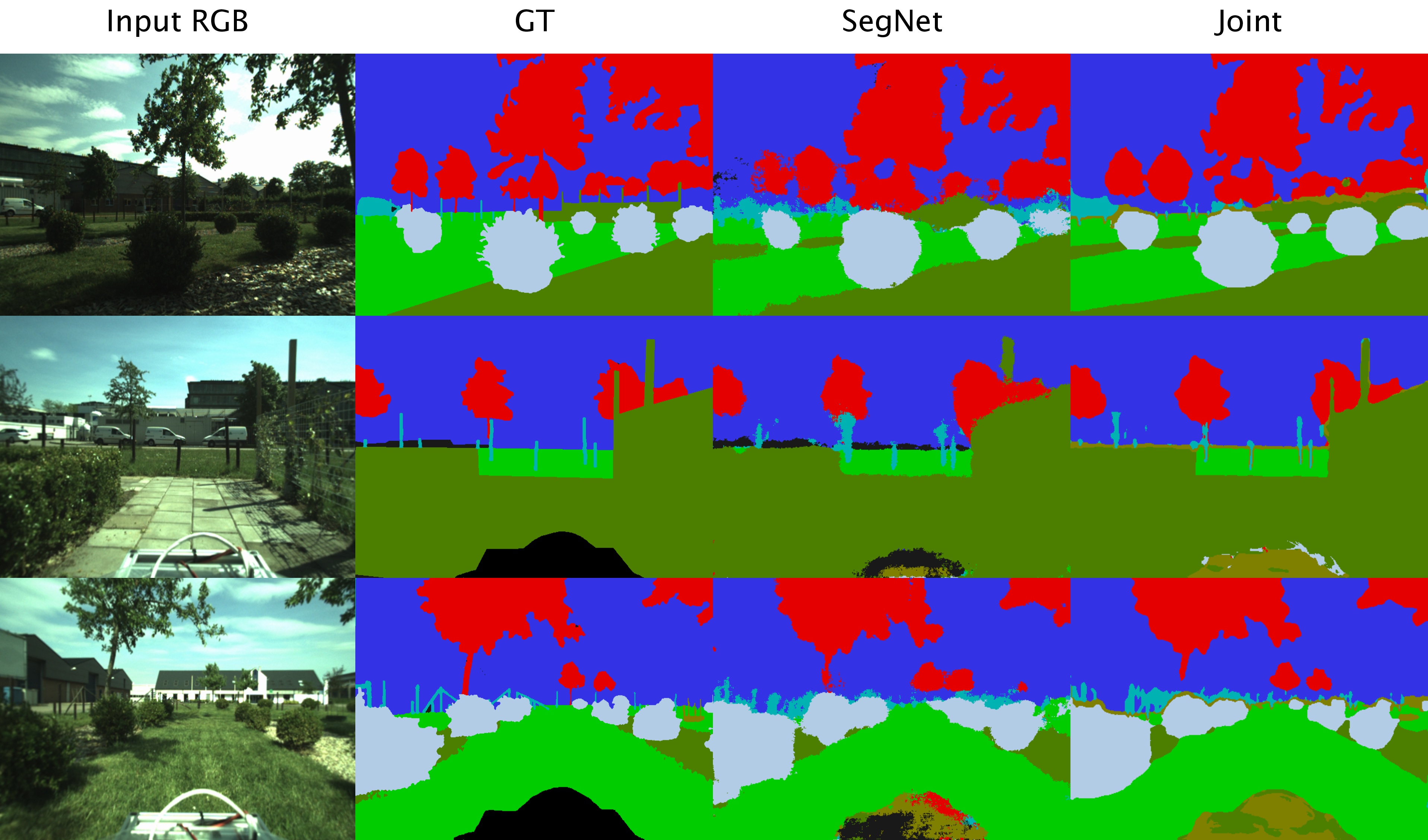}
    \caption{Semantic segmentation evaluation on real world garden images. For segmentation the joint learning performs comparable to the baseline, yet achieves sharper results}
    \label{fig:33}
\end{figure}

In this part, additional result of real world garden images are presented. The proposed method generates reflectance images with more vivid and realistic colors. Moreover, our method mitigates sharp shadow effects better and produces sharper images. Additional results of reflectance images are shown in Fig.~\ref{fig:3}, semantic segmentation results are shown in Fig.~\ref{fig:33}. For segmentation, the joint learning performs comparable to the baseline, yet we achieve sharper results.

\clearpage

\bibliographystyleS{splncs}
\bibliographyS{egbib_supp}

\end{document}